\documentclass{article}

\usepackage{PRIMEarxiv}

\usepackage[utf8]{inputenc} 
\usepackage[T1]{fontenc}    
\usepackage{hyperref}       
\usepackage{url}            
\usepackage{booktabs}       
\usepackage{amsfonts}       
\usepackage{nicefrac}       
\usepackage{microtype}      
\usepackage{lipsum}
\usepackage{fancyhdr}       
\usepackage{graphicx}       
\graphicspath{{./}}     
\usepackage{amsmath}
\usepackage{amssymb}
\usepackage{bm}
\usepackage{graphicx}
\usepackage{longtable}
\usepackage{booktabs}
\usepackage{circuitikz}
\usepackage{indentfirst}
\usepackage{cleveref}
\usepackage[inline,shortlabels]{enumitem}
\usepackage{siunitx}
\usepackage{enumitem}
\usepackage{caption}
\usepackage{subcaption}
\usepackage{booktabs}
\usepackage{multirow}
\usepackage{xcolor}
\usepackage[style=numeric-comp,sorting=none]{biblatex}
\bibliography{references}  

\usepackage{float}

\newlist{E}{enumerate}{1}
\setlist[E]{label=\textbf{E\arabic*:}}
\graphicspath{ {./} }
\crefname{equation}{Eq.}{Eqs.}
\Crefname{equation}{Equation}{Equations}
\crefname{section}{Section}{Sections}
\Crefname{section}{Section}{Sections}
\crefname{figure}{Fig.}{Figs.}      
\Crefname{figure}{Figure}{Figures}  
\crefname{table}{Table}{Tables}
\Crefname{table}{Table}{Tables}

\title{Distributional Reinforcement Learning-based Energy Arbitrage Strategies in Imbalance Settlement Mechanism 
}

\author{
  Seyed Soroush Karimi Madahi \\
  Ghent University \\
  \texttt{seyedsoroush.karimimadahi@ugent.be} \\
  \And
  Bert Claessens \\
  BEEBOP \\
  \And
  Chris Develder \\
  Ghent University \\
}

\begin{document}
\maketitle

\begin{abstract}
Growth in the penetration of renewable energy sources makes supply more uncertain and leads to an increase in the system imbalance. This trend, together with the single imbalance pricing, opens an opportunity for balance responsible parties (BRPs) to perform energy arbitrage in the imbalance settlement mechanism. To this end, we propose a battery control framework based on distributional reinforcement learning (DRL). Our proposed control framework takes a risk-sensitive perspective, allowing BRPs to adjust their risk preferences: we aim to optimize a weighted sum of the arbitrage profit and a risk measure while constraining the daily number of cycles for the battery. We assess the performance of our proposed control framework using the Belgian imbalance prices of 2022 and compare two state-of-the-art RL methods, deep Q learning and soft actor-critic. Results reveal that the distributional soft actor-critic method can outperform other methods. Moreover, we note that our fully risk-averse agent appropriately learns to hedge against the risk related to the unknown imbalance price by (dis)charging the battery only when the agent is more certain about the price.
\end{abstract}

\keywords{Battery energy storage systems (BESS) \and Distributional soft actor-critic (DSAC) \and Imbalance settlement mechanism \and Reinforcement learning (RL) \and Risk-sensitive energy arbitrage}

\section{Introduction}
\label{sec:intro}
Climate change has been a motivation for transitioning toward a decarbonized electricity grid on both the supply and the demand side. The European Commission aims to reach carbon neutrality by 2050~\cite{EU2021}. To achieve this target, the penetration of renewable energy sources (RES) needs to dramatically increase. The International Renewable Energy Agency’s report of 2023 states that the total power capacity of RES in the world grew from 1.57\,TW in 2013 to 3.37\,TW in 2022~\cite{IRENA}. However, this trend makes electricity generation more uncertain due to the dependence of RES production on weather conditions.
Consequently, the increase in the share of RES leads to an increase in the mismatch between generation and consumption.

Given this potentially increasing mismatch between production and consumption, transmission system operators (TSOs) are facing challenges in maintaining the balance of the grid. Following the liberalization of the European electricity system, the balancing responsibility of TSOs has been outsourced to balance responsible parties (BRPs)~\cite{Bottieau2020}. Each unbalanced BRP is penalized by an imbalance price at the end of each imbalance settlement period. According to the electricity balancing guideline (EBGL), published by the European network of transmission system operators for electricity (ENTSO-E), the main objective of the imbalance settlement mechanism is to make sure that BRPs support the system balance in an efficient way and to stimulate market participants in restoring the system balance~\cite{ENTSO}. Also, EBGL states that a single imbalance pricing method should be used to calculate the imbalance cost: the settlement price should be the same for both negative and positive imbalances. Such a single imbalance pricing encourages BRPs to deviate from their day-ahead nomination to help the TSO with balancing the grid and to reduce their cost. The wide usage of RES in addition to the single imbalance pricing provides an opportunity for BRPs to reduce their cost using an arbitrage strategy in the imbalance settlement mechanism. For this purpose, recently battery energy storage systems (BESS) have attracted the attention of BRPs due to their fast response time~\cite{Yang2023}, high efficiency~\cite{Krupp2023}, and significant decreases in cost of recent battery technology~\cite{fida2023optimal}.

Energy arbitrage in the imbalance settlement mechanism is challenging because of high uncertainties in imbalance price and near real-time decision-making.
Due to these mentioned challenges, as well as the recent change in the imbalance pricing methodology, few research works have been conducted on the arbitrage in the imbalance settlement mechanism~\cite{Xiao2020,Bottieau2020,Smets2023,Vejdan2018,Krishnamurthy2017,Alireza2019,Lago2021controller}. Although most of the cited studies have formulated control strategies for BESS using model-based optimization methods (such as stochastic optimization and robust optimization), we argue that model-based optimization methods are not the most appropriate for obtaining an arbitrage strategy. Although it is possible, to formulate the energy arbitrage problem as a nonlinear programming problem, because of the nonconvex nature of such nonlinear problems, there is no efficient way to find the optimal solution for them~\cite{lin2013review}. Hence, linearization techniques (such as piecewise linear approximation) are applied to transform a nonlinear problem into a linear or mixed-integer convex problem. However, applying these techniques may result in an intractable optimization problem or an inaccurate approximation of the problem.
Moreover, these model-based methods need a (probabilistic) forecaster for future imbalance prices to address uncertainty in future prices. In stochastic optimization, such uncertainties can be handled by generating a set of scenarios. Yet, as imbalance prices are highly uncertain, a large number of scenarios are required to correctly reflect the imbalance price distribution, which increases the computational burden to the extent that the problem may become computationally intractable. On the other hand, although robust optimization does not need as many scenarios~\cite{engels2017combined}, its obtained solution might be a very conservative strategy and not necessarily the most economical one~\cite{zhao2023day}.

Given the above challenges, few research works have focused on risk management in the arbitrage problem in the imbalance settlement mechanism. Generally, market participants have different risk preferences. For example, BRPs have more conservative arbitrage strategies in the imbalance settlement mechanism because of highly volatile imbalance prices. In other words, BRPs assign higher weights to scenarios with lower revenues and deviate from risk-neutral decision-making. Thus, to provide a more practical solution, a risk-averse perspective needs to be considered in the arbitrage strategy, while the previous studies have ignored risk management. Moreover, a battery's lifetime mainly depends on its charging/discharging operations. Frequently switching between charging and discharging can significantly reduce the battery cycle life and thus decrease the net profit\, due to an increased operational cost of the BESS. We note that existing works have not investigated the impact of the battery cycle life on the arbitrage strategy.

In summary, shortcomings and weaknesses in previous studies of arbitrage strategies are that:
\begin{enumerate*}[(i),nosep,noitemsep]
    \item adopt model-based methods (which are complex to solve)
    \item do not consider a risk-sensitive perspective; and
    \item neglect battery cycle life constraints.
\end{enumerate*}
To address these shortcomings (further elaborated in \cref{sec:Background}), in this paper, we propose a distributional reinforcement learning (DRL)-based control framework for a risk-sensitive energy arbitrage strategy in the imbalance settlement mechanism for BESS. The proposed control framework (\cref{sec:Problem Formulation}) aims to maximize the arbitrage profit as well as a risk measure by constraining the daily number of cycles for the battery. We believe DRL methods are proper methods for risk management, since they learn the complete probability distribution of random returns instead of the expected return. The proposed control framework can be tuned according to the risk preference of BRPs from a fully risk-averse perspective to a fully risk-seeking one. In this paper, we start from two state-of-the-art reinforcement leaning (RL) methods, i.e., deep Q learning (DQN), as a value-based method, and soft actor-critic (SAC), as a policy gradient method. We extend these vanilla DQN and SAC methods with a distributional perspective (i.e., DDQN, DSAC) and a risk-aware component in the loss function (\cref{sec:RL methods}). The performance of the proposed control framework is evaluated on the Belgian imbalance prices of 2022 (\cref{sec:results,sec:Conclusion}). Overall, our contributions in this paper are that we propose a DRL-based control framework
\begin{enumerate}[(i)]
    \item  for a BESS while considering a constraint on the daily number of cycles;
    \item that achieves a risk-sensitive arbitrage strategy with a tunable risk tolerance by optimizing a weighted sum of the arbitrage revenue and a risk measure in the imbalance settlement mechanism;
    \item for which we compare the performance of value-based and policy gradient RL methods in a highly uncertain trading market.
\end{enumerate}

\section{Background and Related Work}
\label{sec:Background}
Energy arbitrage is a technique to achieve financial profits by purchasing energy when the price is cheap and selling it when the price is expensive~\cite{ellis2023degradation}. Energy arbitrage can be performed within one electricity market (e.g., day-ahead market~\cite{ansari2023bi} or intra-day market~\cite{Boukas2021}) to take advantage of varying prices at different hours. Moreover, energy arbitrage strategies between several electricity markets have been developed to benefit from a price difference between two or more electricity markets, for instance, energy arbitrage between day-ahead and intra-day markets~\cite{Zou2023}, day-ahead and real-time markets~\cite{Alireza2019}, day-ahead market and imbalance settlement mechanism~\cite{ruelens2016}, or day-ahead, intra-day, and real-time markets~\cite{Brijs2019}.

\begin{figure}[h]
    \centering
    \includegraphics[scale=0.49]{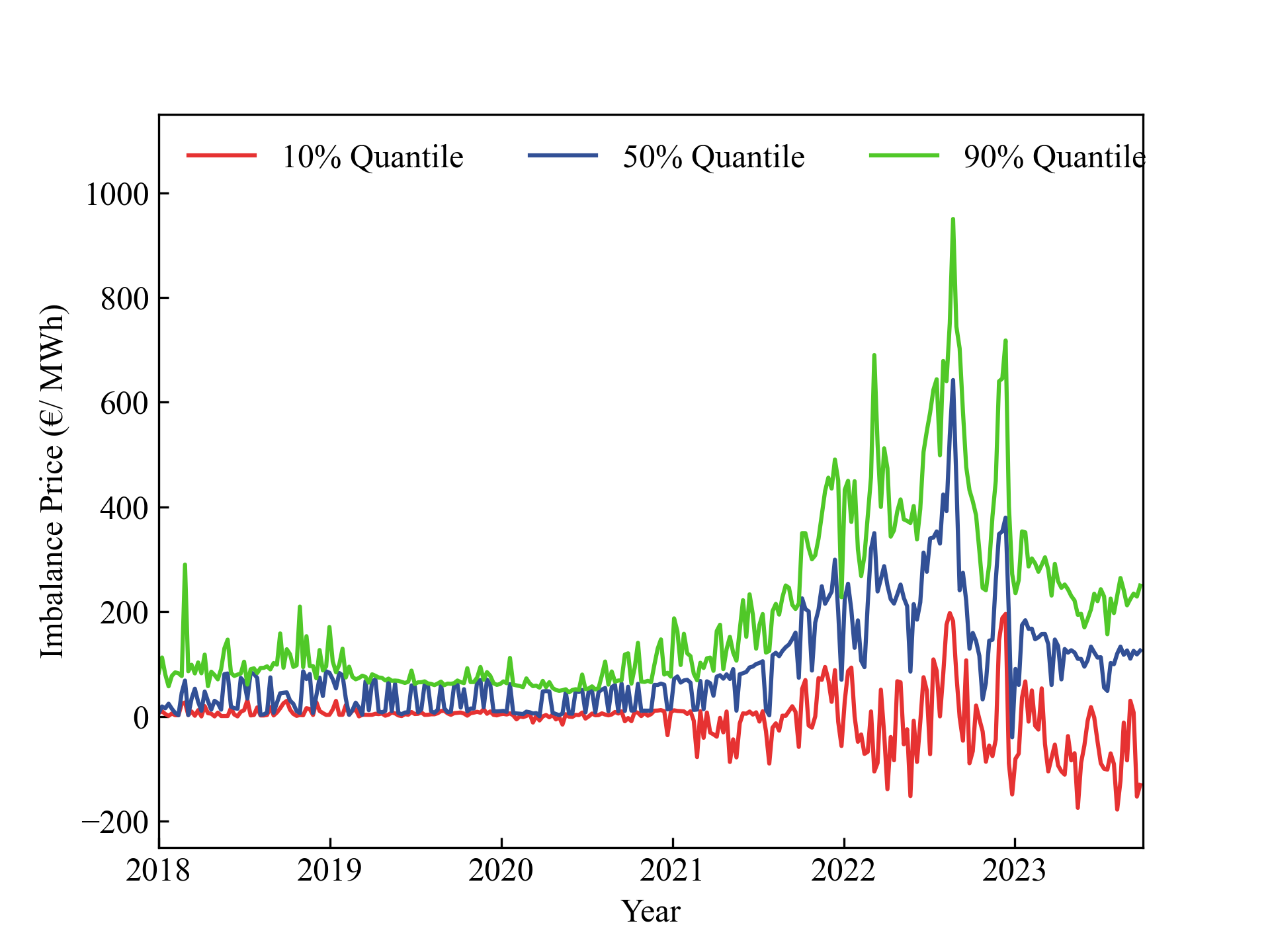}
    \caption{The evolution of Belgian imbalance prices from 2018 to 2023}
    \label{fig:price evolution}
\end{figure}

The recent change in the imbalance price calculation~\cite{ENTSO} and an increase in imbalance prices have opened up a new arbitrage opportunity in electricity markets. \Cref{fig:price evolution} demonstrates the rise in Belgian imbalance prices in recent years. However, only few studies have been conducted on energy arbitrage in the imbalance settlement mechanism, due to the high risk involved in this arbitrage. In~\cite{Xiao2020}, a novel real-time stochastic multi-period management strategy for a virtual power plant was proposed to maximize the profit of a virtual power plant as well as minimize the operational grid cost. It solves a sequential stochastic optimization problem to manage the participation of a BESS in the real-time market. The authors in~\cite{Bottieau2020} first implement a new tailored encoder-decoder architecture to generate improved probabilistic forecasts of the future system imbalance. Then, they solve a bi-level robust optimization problem to maximize the revenue from the participation of a BESS in the imbalance settlement. The authors in~\cite{Smets2023} introduce a novel stochastic model predictive control (MPC) approach to optimize the revenue of BESS in the imbalance settlement mechanism by taking into account battery degradation costs and risk aversion. More specifically, an attention-based recurrent neural network is used to predict the system imbalance and imbalance price. In~\cite{Vejdan2018}, first, the maximum potential profit from the real-time market is obtained using a linear optimization program with the assumption of perfect foresight for future prices. Then, a shrinking-horizon control algorithm is developed to obtain the energy arbitrage strategy for BESS in the real-time market by considering forecast errors in future real-time prices. Reference~\cite{Krishnamurthy2017} proposes a stochastic model to maximize the energy arbitrage revenue of BESS under uncertainty in day-ahead and real-time markets. A hybrid stochastic-robust optimization method is proposed in~\cite{Alireza2019} to maximize the revenue of BESS participants in day-ahead and real-time markets. The day-ahead market problem is solved by stochastic optimization, while the bidding and offering strategy in the real-time market is determined by robust optimization. The authors in~\cite{Lago2021controller} proposes control strategies for seasonal thermal energy storage systems to interact with day-ahead and imbalance markets: MPC-based and RL-based controllers are developed for each market interaction to compare the performance of these two controllers in the different electricity markets. 

Thus, most previous research works have applied \emph{model-based} optimization methods to solve the arbitrage problem~\cite{Xiao2020,Bottieau2020,Smets2023,Vejdan2018,Krishnamurthy2017,Alireza2019}. Nonetheless, the main disadvantage of these methods is that they require linearization techniques to approximate the nonlinear problem as a linear (or mixed-integer) convex problem that can result in an inaccurate approximation. Due to partially known model parameters and uncertainties of the real electricity market, the real market is simplified into a convex market model, resulting in an inaccurate approximation of the real market dynamics~\cite{dolanyi2023capturing}. Furthermore, stochastic optimization can be time-consuming, while robust optimization leads to overly conservative strategies. Another limitation of the mentioned studies is that only one of them \cite{Smets2023} proposes a risk-sensitive arbitrage strategy. Moreover, the previous research works~\cite{Xiao2020,Bottieau2020,Vejdan2018,Krishnamurthy2017,Alireza2019} ignore the effect of a battery's lifetime on the arbitrage strategy.

To avoid problems of model-based optimization methods, \emph{RL methods} can be used. Recently, RL, as a model-free method, has attracted researchers’ attention due to its remarkable performance in solving complex sequential decision-making problems such as playing games, robotic control, and autonomous driving. RL can learn a (near-)optimal policy for a stochastic nonlinear environment by directly interacting with the environment~\cite{Lucian2018}. In RL, there is no special hypothesis regarding the reward function; it can be linear or nonlinear. In contrast to model-based optimization methods, model-free RL methods do not need prior knowledge or an explicit model of the environment. The agent, by interacting with the environment, captures uncertainties and estimates system dynamics. Another advantage of RL methods is that after training the RL agent, its learned policy can be directly used in a new test setting without requiring solving any optimization problem. Therefore, RL methods are efficient tools for real-time control~\cite{Qiu2023}. RL methods have been successfully applied to many power system problems such as the smart charging of EVs~\cite{Sadeghianpourhamami2019,sultanuddin2023development}, demand response~\cite{li2023reinforcement}, frequency control~\cite{yakout2023improved}, etc. Generally, model-free RL methods can be classified into two categories: value-based methods (e.g., Q-learning, fitted Q-iteration (FQI), DQN, etc.) and policy gradient methods (e.g., actor-critic, deep deterministic policy gradient (DDPG), SAC, etc.)~\cite{Cao2020}. In value-based methods, the Q (or V) function is learned (estimated) and the action is chosen based on the learned Q (or V) function as to maximize it. The authors in~\cite{Mnih2015} proposed the DQN method, which combines RL with deep learning. A trained agent using DQN reached human-level performance on many Atari video games. On the other hand, policy gradient methods directly learn the policy. In~\cite{haarnoja2018}, the SAC method has been proposed as an off-policy actor-critic method. In SAC, the policy is learned by an actor network and the Q function is estimated by a critic network. The actor aims to maximize the expected reward as well as the entropy of the actor, to encourage the agent to explore the environment more. In this paper, we will use the DQN (as a state-of-the-art method in value-based methods) and SAC (as a state-of-the-art method in policy gradient methods) methods to solve the arbitrage problem.

\section{Problem Formulation}
\label{sec:Problem Formulation}
In this section, the imbalance settlement mechanism is explained in detail (\cref{sec:imbalance intro}) and the Markov decision process (MDP) formulation of the energy arbitrage problem in the imbalance settlement mechanism is provided (\cref{sec:MDP without cycle,sec:MDP with cycle}).

\subsection{Imbalance Settlement Mechanism}
\label{sec:imbalance intro}
BRPs are responsible for continuously balancing their individual demand and supply. But sometimes BRPs deviate from their traded consumption and generation due to uncertainties in the grid. The total imbalance volume of all BRPs in a single control area is called the total system imbalance~\cite{baetens2020}. Positive and negative values of the system imbalance indicate the excess and shortage of the generation, respectively. A TSO corrects the system imbalance in real-time by activating reserve capacities offered in the balancing market~\cite{Lago2021market}. A TSO charges BRPs for their imbalance at a price specific to the imbalance settlement period (15 min in most European markets). This mechanism is known as imbalance settlement. The imbalance price is dependent on the reserve volume activated by the TSO~\cite{vatandoust2023}.  In each imbalance settlement period, the negative imbalance price is equal to the highest activated upward reserve offer (marginal incremental price), and the positive imbalance price is determined by the lowest activated downward reserve offer (marginal decremental price)~\cite{Marneris2022}. Three main imbalance pricing methodologies are used in various countries: (1) dual pricing; (2) two-price settlement; and (3) single pricing~\cite{Marneris2022}.

In the dual pricing method, the imbalance price is different for positive and negative imbalances. BRPs penalize for negative and positive imbalances using the marginal incremental price (MIP) and marginal decremental price (MDP), respectively. This pricing method motivates BRPs to keep the balance within their own portfolio without being concerned about the total system imbalance. The main drawback of this method is that there is no incentive for BRPs to deviate from their nomination to restore the grid. For instance, if the total system imbalance is positive and there is a BRP that can reduce this imbalance, then this BRP is not incentivized, but even penalized for deviating from its day-ahead nomination.

In the two-price settlement method, similar to the dual pricing method, different imbalance prices are considered for each imbalance direction. The difference with the dual pricing method is that if the imbalance direction of BRPs is opposite to the total system imbalance direction, the imbalance price is the same as the day-ahead price. Although in this pricing method, BRPs do not face penalties due to their deviation for helping TSO with restoring the grid, the imbalance price is not attractive to create a portfolio imbalance for supporting the grid (typically, day ahead prices are imbalance prices).

In the single pricing method, the imbalance price is the same for both imbalance directions and depends on the total system imbalance. This pricing method provides an opportunity for BRPs to reduce their cost by supporting the grid. For instance, if the total imbalance price is negative and a BRP creates a positive imbalance, the BRP will receive an MIP (imbalance price) which is usually higher than the day-ahead price. In some countries, e.g., Germany, despite using the single pricing method, arbitrage in the imbalance settlement mechanism is prohibited and market players are expected to trade honestly in the markets~\cite{Matsumoto2021}. Nonetheless, the arbitrage in the imbalance settlement mechanism is a win-win situation for both BRPs and TSOs. On the one hand, BRPs can profit from the arbitrage and indirectly reduce the total system imbalance. On the other hand, this decrease in the total system imbalance results in a lower imbalance price since the TSO does not need to activate more expensive reserve offers.

As mentioned earlier, ENTSO-E aims to harmonize the imbalance settlement mechanism in Europe by implementing the single pricing method for calculating the imbalance price with a 15 minute imbalance settlement period. For this reason, the focus of this paper is on the single pricing method. The Belgian imbalance settlement mechanism is a good case study for this research work because since the beginning of 2020, it adopts the single pricing method with a 15 minute settlement period~\cite{baetens2020}.

\subsection{MDP Formulation without Cycle Constraint Consideration}
\label{sec:MDP without cycle}
The energy arbitrage problem can be formulated as an MDP. An MDP provides a mathematical framework for stochastic sequential decision-making problems and is modeled by a tuple $(\mathcal{S},\mathcal{A},\mathcal{R},\mathcal{P},\gamma)$, where $\mathcal{S}$ is the state space, $\mathcal{A}$ is the (discrete) action space, $\mathcal{R}: \mathcal{S}\times\mathcal{A}\rightarrow\mathbb{R}$ represents the immediate reward function, $\mathcal{P}: \mathcal{S}\times\mathcal{S}\times\mathcal{A}\rightarrow[0,1]$ denotes the unknown state transition probability distribution, and $\gamma \in (0,1]$ is the discount factor~\cite{Sutton2018}. At each time step $t$, the agent observes the environment state $s_t \in \mathcal{S}$ and takes an action $a_t \in \mathcal{A}$ based on the current state. As a consequence of the taken action, the agent receives a reward value  $\mathcal{R}(s_t,a_t)$ and moves to a new state $s_\text{t+1} \in \mathcal{S}$ with the probability determined by the state transition probability distribution $\mathcal{P}(s_\text{t+1}|s_t,a_t)$. In the energy arbitrage problem, the agent is a decision maker who decides about the charging/discharging of BESS at each time step. The environment is the external context with which the agent interacts (electricity markets, grid, etc.). We define the MDP formulation of the energy arbitrage problem in the imbalance settlement mechanism without cycle constraints as follows:
\begin{enumerate}[(i)]
\item \emph{State}: The state at each time step is expressed as
\begin{equation}
    s_t=(T_\textrm{qh},\textrm{qh},\textrm{mo},\textrm{SOC}_t,\hat{\pi}^\textrm{imb}_t)
    \label{1}
\end{equation}
where $T_\textrm{qh}$ represents the minute of the quarter hour, $\textrm{qh}$ is the quarter hour of the day, $\textrm{mo}$ is the month of the year, $\textrm{SOC}_t$ is the SoC of BESS at time $t$, and $\hat{\pi}^\textrm{imb}_t$ is the forecasted imbalance price of the current quarter-hour. We used a forecast of the imbalance price because the real imbalance price of the quarter hour is only calculated once the quarter hour is over.

\item \emph{Action}: We consider a discrete action space with 3 possible actions, as follows:
\begin{equation}
    a_t \in \mathcal{A}, \quad \mathcal{A}=\{-P_\text{max}, 0, P_\text{max}\}
    \label{2}
\end{equation}
where  $P_\text{max}$ is the maximum (dis-)charging power of the BESS. The action $a_t$ represents a decision on the charging/discharging power at time $t$.
\item \emph{Reward}: The objective of the agent is to maximize the revenue by buying energy when the imbalance price is low and selling it when the imbalance price is high. Hence, the reward function to be maximized is the negative of the energy cost, defined as follows
\begin{equation}
    r_t=-a_t\pi^\text{imb}_t
    \label{3}
\end{equation}
where $\pi^\text{imb}_t$ is the real imbalance price of the quarter hour in which $t$ lies.
\item \emph{State transition function}: In the MDP framework, system dynamics are described by a state transition probability function $\mathcal{P}$. This probability function is unknown in the energy arbitrage problem because of uncertainties in the imbalance price. The agent strives to estimate the state probability distribution through interactions with the environment. However, the state transition for $\textrm{SOC}_t$ is controlled by $a_t$ and can be explicitly formulated as below.
\begin{equation}
    \textrm{SOC}_\text{t+1} = \begin{cases}
        \textrm{SOC}^\text{temp}_\text{t+1} &: 0<\textrm{SOC}^\text{temp}_\text{t+1}<1 \\
        0 &: \textrm{SOC}^\text{temp}_\text{t+1}<0 \\
        1 &: \textrm{SOC}^\text{temp}_\text{t+1}>1
    \end{cases}
    \label{4}
\end{equation}
\begin{equation}
    \textrm{SOC}^\text{temp}_\text{t+1}=\textrm{SOC}_t+(\max(a_t,0) \eta_\text{cha}+\frac{\min(a_t,0)}{\eta_\text{dis}})\frac{\Delta t}{C_\text{BESS}}
    \label{5}
\end{equation}
where $C_\text{BESS}$ is the maximum capacity of the BESS, and $\eta_\text{cha}$ and $\eta_\text{dis}$, denote the charging and discharging efficiency of the BESS, respectively.
\end{enumerate}

\subsection{MDP Formulation with Cycle Constraint Consideration}
\label{sec:MDP with cycle}
Frequent charging/discharging cycles cause an extra cost because they expedite the degradation of BESS. Modeling the aging of BESS is crucial as it indicates a capital loss of BESS investment costs~\cite{Engels2019}. Due to the dependence of battery lifetime on its operational strategy, the lifetime of a BESS plays an important role in the financial evaluation of the energy arbitrage strategy. Usually, the lifetime of a BESS is determined by the number of complete charge-discharge cycles before its nominal capacity becomes lower than a certain level of its initial rated capacity~\cite{Zhou2011}. Thus, we constrain the daily number of cycles, since it aligns with the designed lifetime and guarantee provided by manufacturers~\cite{Hu2022}. The MDP formulation with cycle constraint consideration is described next.
\begin{enumerate}[(i)]
    \item \emph{State}: The state is given by
    \begin{equation}
        s_t=(T_\textrm{qh},\textrm{qh},\textrm{mo},\textrm{SOC}_t,\hat{\pi}^\text{imb}_t,n_t^\text{cyc})
    \label{6}
    \end{equation}
    \begin{equation}
        n_t^\text{cyc}=\sum_\text{i=0}^\text{t-1} \frac{|\min(a_i,0)| \Delta t}{C_\text{BESS}}
        \label{7}
    \end{equation}
    where $n_t^\text{cyc}$ is the daily consumed number of cycles, calculated using \eqref{7}.
    \item \emph{Action}: Similar to the MDP formulation without cycle constraints, the action space is discrete with 3 possible actions. The action is determined as follows
    \begin{equation}
        a_t=B(u_t,n_t^\text{cyc}), \quad u_t \in \mathcal{A}=\{-P_\text{max}, 0, P_\text{max}\}
        \label{8}
    \end{equation}
    \begin{equation}
        B(u_t,n_t^\text{cyc})= \begin{cases}
            0 &: u_t<0 \wedge n_t^\text{cyc}>n_\text{max}^\text{cyc} \\
            u_t &: else
        \end{cases}
        \label{9}
    \end{equation}
where  $n_\text{max}^\text{cyc}$ is the maximum allowed daily number of cycles and $B(.)$ is a backup controller to ensure the daily cycle constraint. The backup controller is used to override the agent action ($u_t$) when the agent wants to discharge the battery and the daily number of cycles exceeds the maximum allowed value.
\item \emph{Reward}: The reward function definition is the same as that of the MDP formulation without cycle constraint.
\item \emph{State transition function}: Also the state transition function is the same as that of the MDP formulation without cycle constraint.
\end{enumerate}

\section{Reinforcement Learning Methods}
\label{sec:RL methods}
In this paper, RL methods are used to solve the arbitrage problem formulated as an MDP and find an arbitrage strategy in the imbalance settlement mechanism. The goal in RL is to learn a policy that maximizes the expected long-term reward. Next we detail the two RL methods adopted in this paper, i.e., DQN and SAC. Subsequently, we introduce the distributional perspective on RL and the risk-sensitive RL framework.
\subsection{DQN}
Classical tabular RL methods, e.g., Q-learning, suffer from an issue known as the curse of dimensionality. Since these methods can only be applied to problems with discrete state space, they can not be used for problems with high-dimensional or continuous state space. In addition, these methods usually need handcrafted state representations~\cite{Cao2020}. To overcome these limitations, the DQN method uses a deep neural network as a function approximator to estimate the Q-value function parametrized by $\theta$. The Q-value function $Q_\theta(s_t,a_t)$ is learned by minimizing the following loss function:
\begin{equation}
    L_Q(\theta)=\mathbb{E}_{(s_t,a_t,r_t,s_\text{t+1}) \sim \mathcal{D}} \left[(r_t+\gamma \max_a Q_\text{$\theta'$}(s_\text{t+1},a)-Q_\theta(s_t,a_t))^2 \right].
    \label{10}
\end{equation}
The first benefit of DQN is its stability in learning. In~\cite{Mnih2015}, two techniques are used to stabilize the learning process. First, the target Q function $Q_\text{$\theta'$}(s_t,a_t)$ is used to calculate next state-action values in \cref{10}. Parameters of the target Q function ($\theta'$) are periodically updated with the most recent $\theta$. Second, agent past transitions are stored in an experience replay buffer ($\mathcal{D}$) and for training $Q_\theta$, mini-batches of experiences are sampled from $\mathcal{D}$. Another benefit of the DQN method is that this method is an off-policy method. The key advantage of off-policy methods is their capacity to learn from historical data since using the current experiences as the training set can easily overfit the policy because the training samples are not independent~\cite{Srivastava2014}. In an off-policy setting, a policy learned by the agent is different from a behavior policy used for collecting historical data. Using past transitions for training can significantly improve sample efficiency.
\subsection{SAC}
Value-based methods have some limitations. The application of these methods is limited to problems with a discrete and low-dimensional action space. Also, these methods learn a deterministic policy, which means for a given state, an action taken by the agent is always the same. Thus, keeping a balance between exploration and exploitation in value-based methods is challenging. Policy gradient methods solve these limitations by learning a policy network that outputs the probability of taking actions in each state. From the existing policy gradient methods, we use SAC because of its superior sample efficiency and stability. In this off-policy method, the policy is learned by an actor network $\pi_\phi$ and the Q function is approximated by a critic network $Q_\theta$. The objective of the actor is to maximize the expected reward as well as maximize the entropy of the actor to encourage the agent to further explore the environment. The loss function of the actor network ($J_\pi$) is given by
\begin{equation}
    J_\pi(\phi)=\mathbb{E}_{s \sim \mathcal{D}, a \sim \pi_\phi}[\alpha \ln \pi_\phi(a|s)-Q_\theta(s,a)]
    \label{11}
\end{equation}
The critic network estimates the soft Q-value function. The loss function of the critic network ($L_Q$) is formulated as follows:
\begin{equation}
    L_Q(\theta)=\mathbb{E}_{(s_t,a_t) \sim \mathcal{D}} [(y_t-Q_\theta(s_t,a_t))^2]
    \label{12}
\end{equation}
\begin{equation}
    y_t=r_t+\gamma \mathbb{E}_{a_\text{t+1} \sim \pi_\phi} [Q_\text{$\theta'$}(s_\text{t+1},a_\text{t+1})-\alpha \ln \pi_\phi(a_\text{t+1}|s_\text{t+1})]
    \label{13}
\end{equation}
\begin{equation}
    \theta'=\tau \theta + (1-\tau) \theta'
    \label{14}
\end{equation}
In \cref{13}, $y_t$ is an estimated soft-Q value that is calculated by a modified Bellman equation (the so-called soft Bellman equation). Similar to the DQN method, the target Q function is used to calculate $y_t$. After each update of $Q_\theta$, the parameters of $Q_\text{$\theta'$}$ are updated according to \cref{14} with $\tau \ll 1$ to slowly track the learned network~\cite{Lillicrap2015}.
\subsection{Distributional RL}
A distributional perspective on RL was first introduced in~\cite{bellemare2017}. In DRL methods, the probability distribution over returns is estimated rather than a point estimate of the mean. DRL methods offer several advantages, including more stable learning~\cite{bellemare2017}, mitigating Q-value overestimation~\cite{Duan2021}, and providing a framework for risk-sensitive learning~\cite{Thibaut2023}.
In the vanilla DQN method, the core idea is to estimate the Q-value function $Q_\theta$. Going beyond the vanilla DQN method, the distributional DQN (DDQN) method learns the probability distribution of returns ($\mathcal{Z}_\theta$) using the distributional Bellman equation as follows~\cite{bellemare2017}:
\begin{equation}
L_\mathcal{Z}(\theta)=\mathbb{E}_{(s_t,a_t) \sim \mathcal{D}}[D_\text{KL}(\mathcal{T} \mathcal{Z}_\text{$\theta'$}(s_t,a_t)||\mathcal{Z}_\theta(s_t,a_t))]
\label{15}
\end{equation}
\begin{equation}
\mathcal{T}Z(s_t,a_t)\overset{D}{=}r_t+\gamma \max_a \mathbb{E}_{Z \sim \mathcal{Z}_\text{$\theta'$}}[Z(s_\text{r+1},a)]
    \label{16}
\end{equation}
where $\mathcal{Z}$ is the distribution of returns, $A\overset{D}{=}B$ denotes that two random variables $A$ and $B$ have an equal probability distribution, and $\mathcal{T}\mathcal{Z}_\theta$ indicates the probability distribution of $\mathcal{T}Z$. The distribution of returns can be modeled as a categorical distribution as below.
\begin{equation}
Z(s_t,a_t)= \left\{ z_i \Big| z_i=V_\text{min}+\frac{V_\text{max}-V_\text{min}}{N-1}i, 0 \leq i < N\right \}
    \label{17}
\end{equation}
In \cref{17}, $V_\text{min}$ and $V_\text{max}$ are the maximum and minimum values of random returns, respectively, and $N$ is the number of bins.
In distributional SAC (DSAC), the critic network learns the probability distribution of soft returns. The loss function of the critic network in DSAC is similar to \cref{15}, but the calculation of $\mathcal{T}Z(s_t,a_t)$ differs as follows:
\begin{equation}
   \mathcal{T}Z(s_t,a_t)\overset{D}{=}r_t+\gamma \mathbb{E}_{a_\text{t+1} \sim \pi_\phi, Z \sim \mathcal{Z}_\text{$\theta'$}} \left[Z(s_\text{t+1},a_\text{t+1})-\alpha \ln \pi_\phi(a_\text{t+1}|s_\text{t+1})\right]
    \label{18}
\end{equation}
Since the expectation of $Z(s_t,a_t )$ over $\mathcal{Z}_\theta $ is equal to $Q(s_t,a_t )$, the loss function of the actor network is modified as below.
\begin{equation}
    J_\pi(\phi)=\mathbb{E}_{s \sim \mathcal{D}, a \sim \pi_\phi}[\alpha \ln \pi_\phi(a|s)-\mathbb{E}_{Z \sim \mathcal{Z}_\text{$\theta$}}[Z(s,a)]]
    \label{19}
\end{equation}
\subsection{Risk-sensitive RL}
By approximating the probability distribution of returns, DRL presents a possibility for learning a risk-averse policy. In a risk-neutral RL framework, the agent in each state takes an action that aims to maximize the expected return (Q value). On the other hand, in the risk-sensitive RL framework, the agent takes an action with the lowest associated risk. The main risk in the arbitrage problem is related to forecasted imbalance prices. The greater the inaccuracy in predicted prices, the higher the associated risk of taking the wrong action.

Risk measures can be used to assess the level of risk associated with a distribution of returns~\cite{Ma2020}. The loss function of the actor network in the risk-sensitive DSAC can be formulated as follows:
\begin{equation}
    J_\pi(\phi)=\mathbb{E}_{s \sim \mathcal{D}, a \sim \pi_\phi} \left[\,\alpha \ln \pi_\phi(a|s)- \mathbb{E}_{Z \sim \mathcal{Z}_\text{$\theta$}}[Z(s,a)] - \beta \Psi [Z(s,a)]\, \right],
    \label{20}
\end{equation}

where $\Psi[.]$ represents a risk measure function and $\beta$ is a parameter that controls the trade-off between the expectation value and risk. $\beta=0$ represents the risk-neutral attitude of the agent. As $\beta$ increases, the agent becomes more risk-averse. In this paper, value-at-risk (VaR) is applied as the risk measure function:
\begin{equation}
    \textrm{VaR}_\rho (Z) = \inf \{ z|\textrm{CDF}_Z(z) \geq \rho \},
    \label{22}
\end{equation}
where $\rho \in (0,1]$ is a confidence level. We will set $\rho=0.1$ in this paper.

\section{Simulation Results}
\label{sec:results}
 We will evaluate the performance of the proposed control framework, is explained in \cref{sec:Problem Formulation,,sec:RL methods}, for the energy arbitrage problem.

\subsection{Experimental Setup}
\begin{figure}[h]
    \centering
    \includegraphics[scale=0.23]{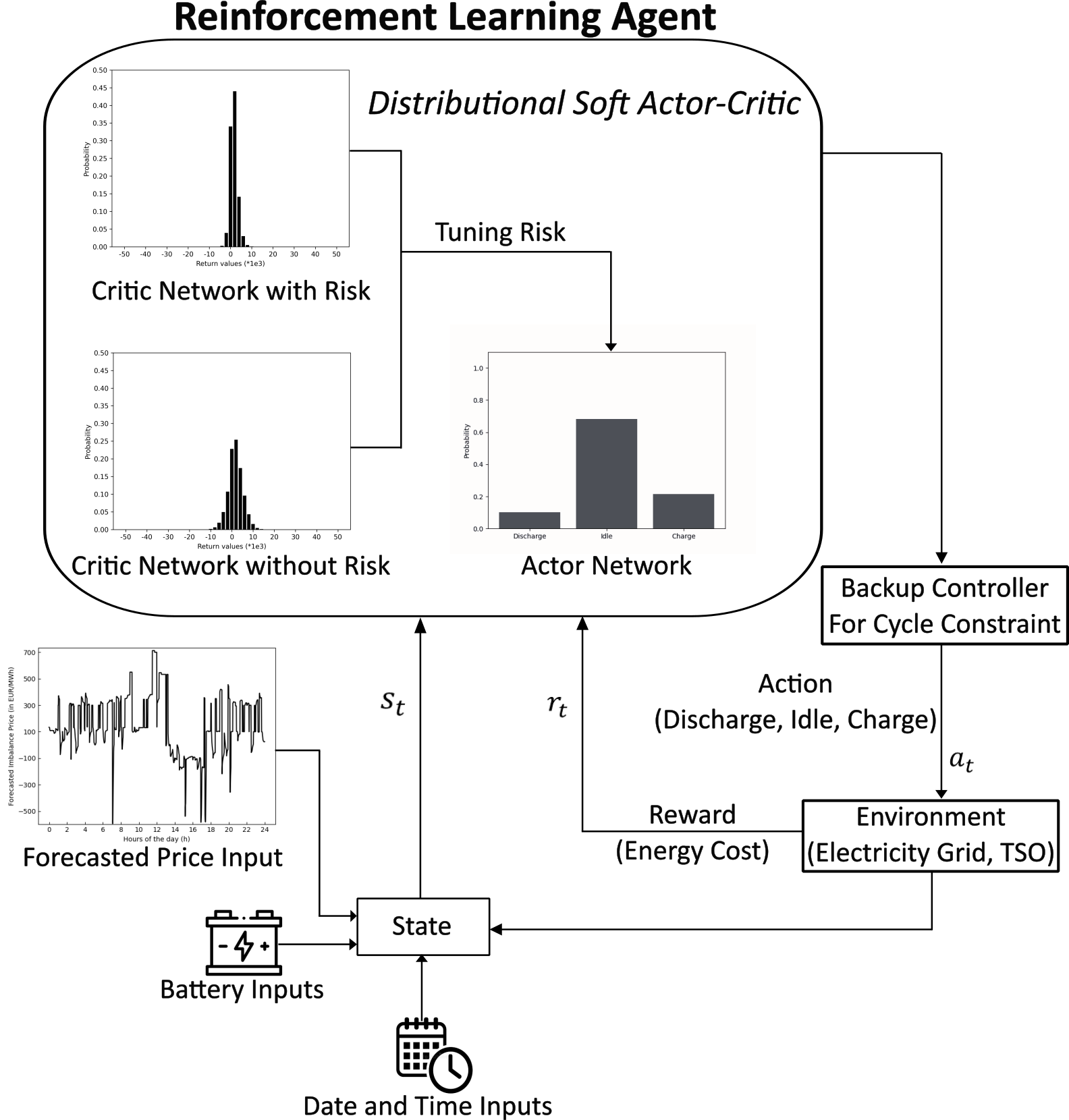}
    \caption{The overview of the proposed control framework}
    \label{fig:overview}
\end{figure}

\Cref{fig:overview} shows the overview of the proposed control framework, which we test on the Belgian imbalance in 2022 extracted from~\cite{Elia}. As mentioned in \cref{sec:imbalance intro}, Elia publishes two imbalance prices: 15-minute-based and 1-minute-based prices. The reference price for the imbalance settlement of BRPs is the 15-minute-based price which is the real imbalance price calculated at the end of the quarter-hour period. The 1-minute-based prices, on the other hand, are calculated based on non-validated data, based on the instantaneous system imbalance and prices of cumulative activated regulation volumes on a minute basis. These 1-minute-based prices are published to provide additional information to BRPs~\cite{Elia2019}. We use these non-validated prices as forecasted imbalance prices of the corresponding quarter-hour period. Since the granularity of the forecasted imbalance prices is one minute, the RL agent takes an action every minute. In this work, the day-ahead schedule for the battery is set to zero which means that the battery does not trade in the day-ahead market. However, future work will extend our proposed control framework for arbitrage in both the day-ahead market and imbalance settlement. To train and validate the proposed control framework, the imbalance price dataset is split as follows: the first 20 days of each month are considered as the training set, the 21st day to the 25th day of each month are considered as the validation set, and the remaining days of each month are used as the test set. The considered BESS has a power rating of 1 MW and a maximum capacity of 2 MWh with a round-trip efficiency of 0.9 for both charging and discharging. Since the maximum allowed annual number of cycles for the BESS is 400, the maximum daily number of cycles is set to 1.1. The RL methods are trained with \num{50000} episodes and each episode constitutes a simple day. The hyperparameters used for the methods are listed in \cref{tab:hyperparameters}. The proposed control framework is implemented in Python using the PyTorch package.

We design experiments to answer the following questions:
\begin{itemize}
    \item Q1: What is the learned arbitrage strategy when there is no limit on the daily number of cycles? 
    \item Q2: How does a daily number of cycles affect the learned arbitrage strategy?
    \item Q3: What is the effect of the risk-averse perspective on the learned arbitrage strategy?
\end{itemize}

\begin{table}[h]
    \centering
    \caption{Method hyperparameters}
    {\begin{tabular}{cccccc}
        \toprule
        \multicolumn{2}{c}{Shared} & \multicolumn{2}{c}{DQN} & \multicolumn{2}{c}{SAC} \\
        \cmidrule(lr){1-2} \cmidrule(lr){3-4} \cmidrule(lr){5-6}
        {Parameter} & {Value} & {Parameter} & {Value} & {Parameter} & {Value} \\
        \midrule
        Discount factor $\gamma$ & 0.9995  & learning rate & \num{5e-4} & actor learning rate & \num{2e-5}\\
        Soft update factor $\tau$ & 0.1 &  &  & critic learning rate & \num{1e-4}\\
        Experience buffer size & \num{1e6} &  &  & initial $\alpha$ & 1\\
        Mini-batch size & \num{16384} &  &  & $\alpha$ learning rate & \num{3e-4}\\
        Network hidden layer size & [256,128] &  &  &  &\\
        $V_\text{max}$ & \num{5000} &  &  &  & \\
        $V_\text{min}$ & $-$\num{5000} &  &  &  & \\
        $N$ & 11 &  &  &  & \\
        \bottomrule
    \end{tabular}}
    \label{tab:hyperparameters}
\end{table}

\subsection{Arbitrage Strategy without Cycle Constraint (Q1)}
\begin{figure}[H]
    \centering
    \begin{subfigure}{0.49\textwidth}
        \includegraphics[scale=0.5]{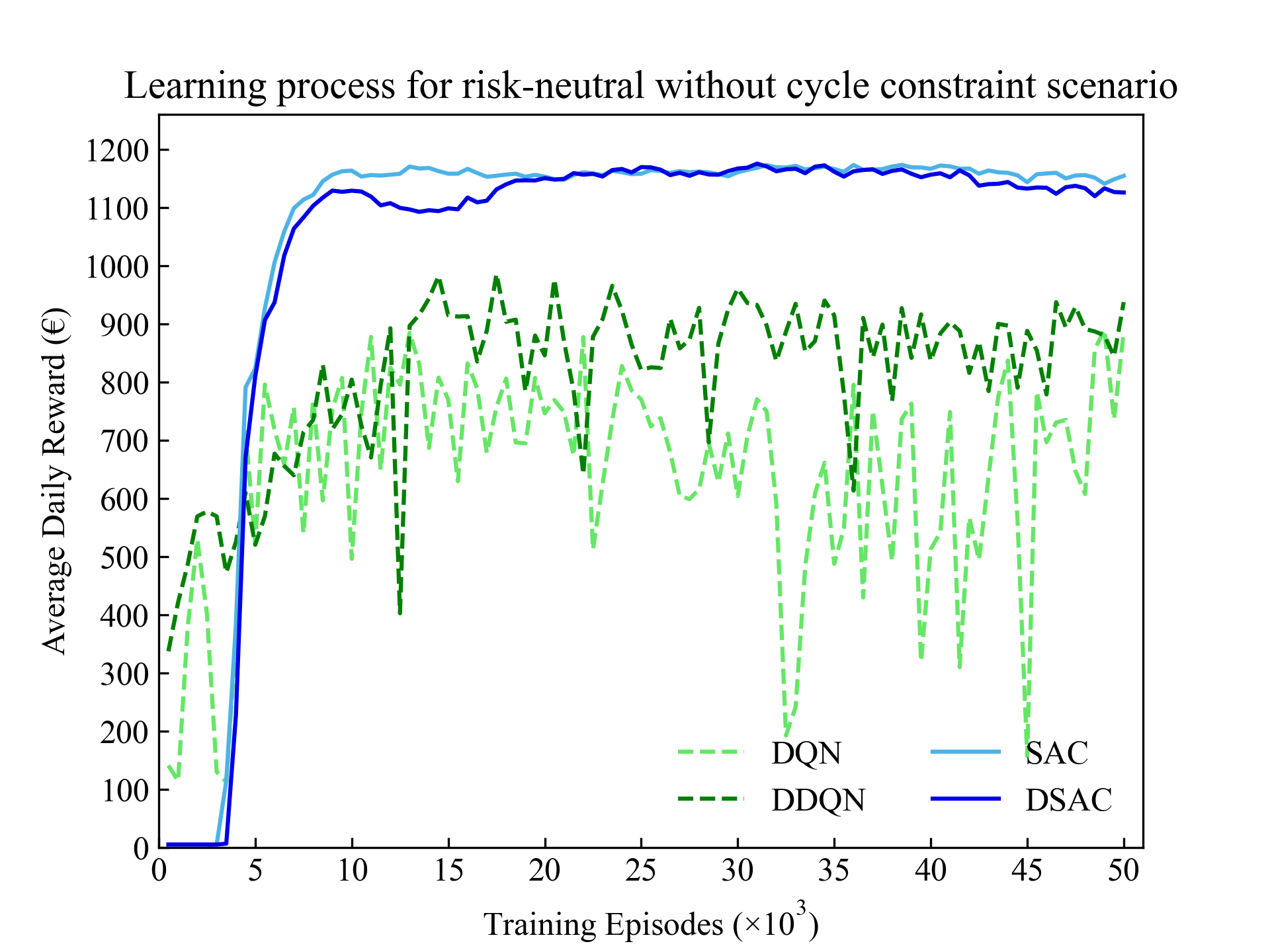}
        \caption{}
        \label{fig:learning process without cycle_reward}
    \end{subfigure}
    \hfill
    \begin{subfigure}{0.49\textwidth}
        \includegraphics[scale=0.5]{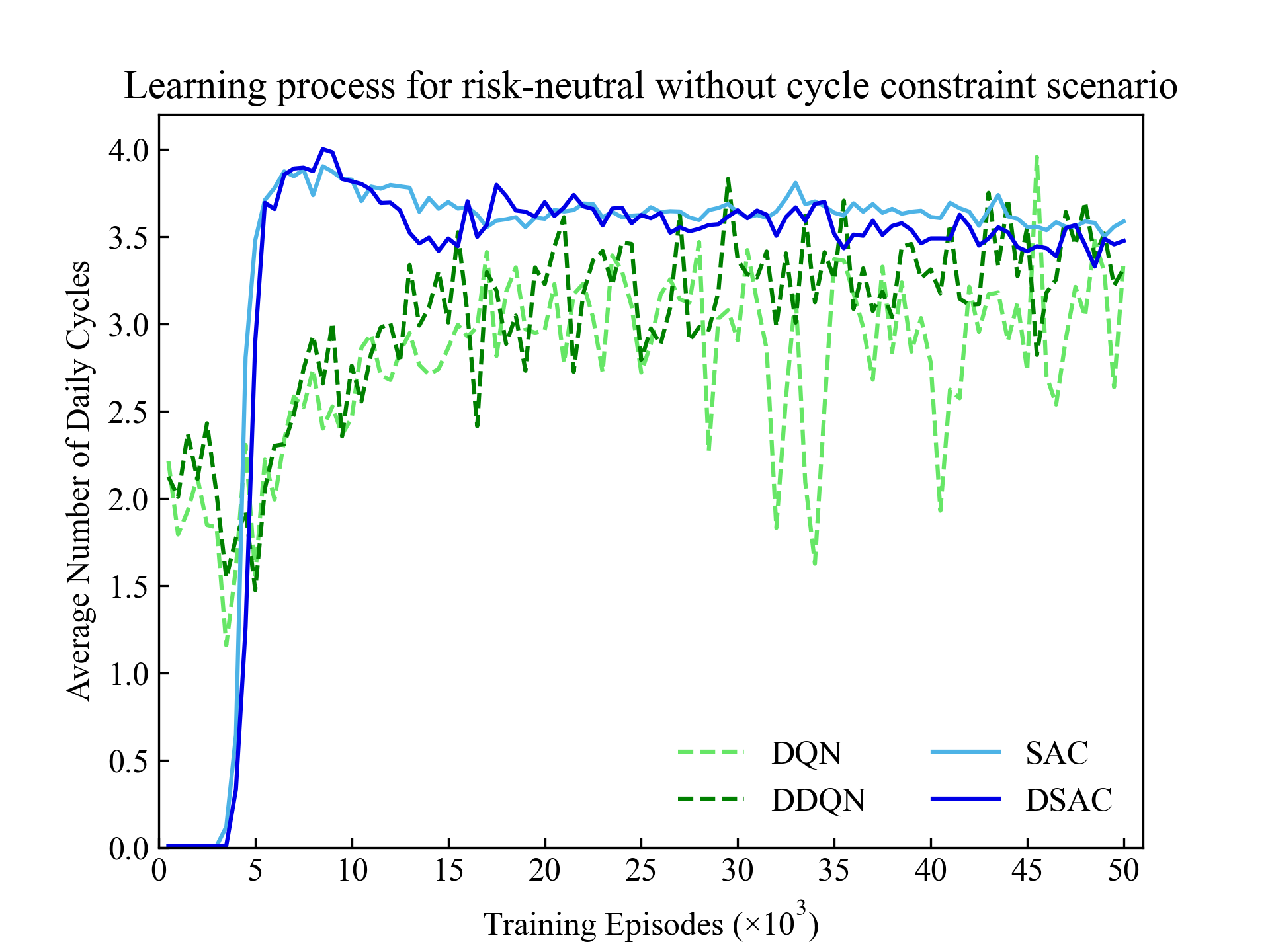}
        \caption{}
        \label{fig:learning process without cycle_cycle}
    \end{subfigure}
    \caption{The learning process of the four RL methods for the risk-neutral without cycle constraint scenario. (a) The average daily profit of the RL methods. (b) The average daily number of cycles.}
    \label{fig:learning process without cycle}
\end{figure}

The learning process of the RL methods for the risk-neutral scenario, without considering the cycle constraint, is illustrated in \cref{fig:learning process without cycle}. The performance of the trained RL methods on the test set is indicated in \cref{tab:risk-neutral results}. Results show that the DRL methods outperform the standard RL methods. The reason behind this is that estimating the probability distribution of returns, rather than the expectation of returns, can provide a more stable training target. Also, the DRL methods can mitigate instability in the Bellman optimality operator by learning probability distribution of returns. The DDQN method increases the average daily profit by 17\% compared to the DQN method. DSAC improves the proportional reward (defined as the ratio of average daily profit to average daily number of cycles) by 2.1\% compared to SAC. The comparison between the performance of the distributional and vanilla DQN, and SAC, indicates that the distributional perspective can enhance DQN results to a greater extent. The reason is that the SAC method mitigates instability in the Bellman optimality operator by using an actor network instead of the max operator in the Bellman equation. Therefore, the improvement in the DSAC results is mainly due to stable training target for the critic network. However, the distributional perspective can boost the performance of the vanilla DQN by both providing stable training targets and mitigating instability in the Bellman optimality operator. Results also highlight the superiority of SAC over DQN. This is because SAC can mitigate Q-value overestimations in DQN by replacing the max operator (\cref{10}) with the expectation operator (\cref{13}) in the Bellman equation.

\begin{table}[H]
    \centering
    \caption{Evaluation of RL methods on the test set in the risk-neutral scenarios}
    \begin{tabular}{ccccccc}
        \toprule
        \multirow{4}{*}{Methods} & \multicolumn{3}{c}{without cycle constraint} & \multicolumn{3}{c}{with cycle constraint}\\
        \cmidrule(lr){2-4} \cmidrule(lr){5-7} 
        {} & \begin{tabular}{@{}c@{}} Profit \\ (\texteuro/ per day)\end{tabular} & \begin{tabular}{@{}c@{}} Cycles \\ (per day)\end{tabular} & \begin{tabular}{@{}c@{}} Proportional profit \\ (\texteuro/ per cycle)\end{tabular}  & \begin{tabular}{@{}c@{}} Profit \\ (\texteuro/ per day)\end{tabular} & \begin{tabular}{@{}c@{}} Cycles \\ (per day)\end{tabular} & \begin{tabular}{@{}c@{}} Proportional profit \\ (\texteuro/ per cycle)\end{tabular} \\
        \midrule
        DQN & 749.9  & 3.2 & 235.6 & 338.0 & 0.9 & 399.1 \\
        DDQN & 877.5 & 3.2 & 275.9 & 397.2 & 1 & 405.9 \\
        SAC & \num{1147.6} & 3.7 & 307.6 & 504.9 & 1.1 & 472.7 \\
        DSAC & \num{1148.5} & 3.6 & 314.1 & 486.4 & 0.9 & 541.7 \\
        \bottomrule
    \end{tabular}
    \label{tab:risk-neutral results}
\end{table}

\begin{figure}[H]
    \centering
    \begin{subfigure}{0.49\textwidth}
        \includegraphics[width=\textwidth]{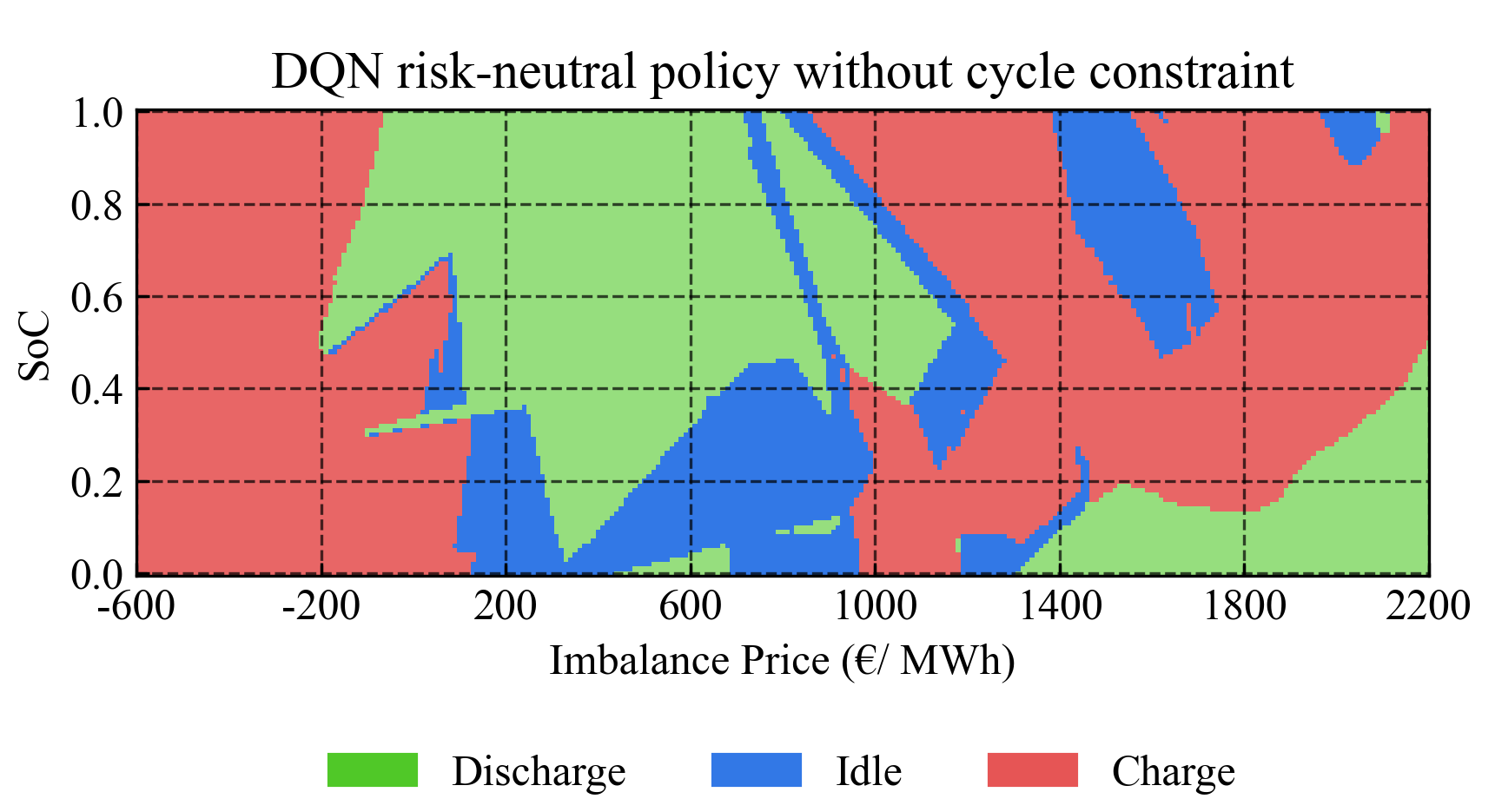}
        \caption{}
        \label{fig:DQN policy without cycle}
    \end{subfigure}
    \hfill
    \begin{subfigure}{0.49\textwidth}
        \includegraphics[width=\textwidth]{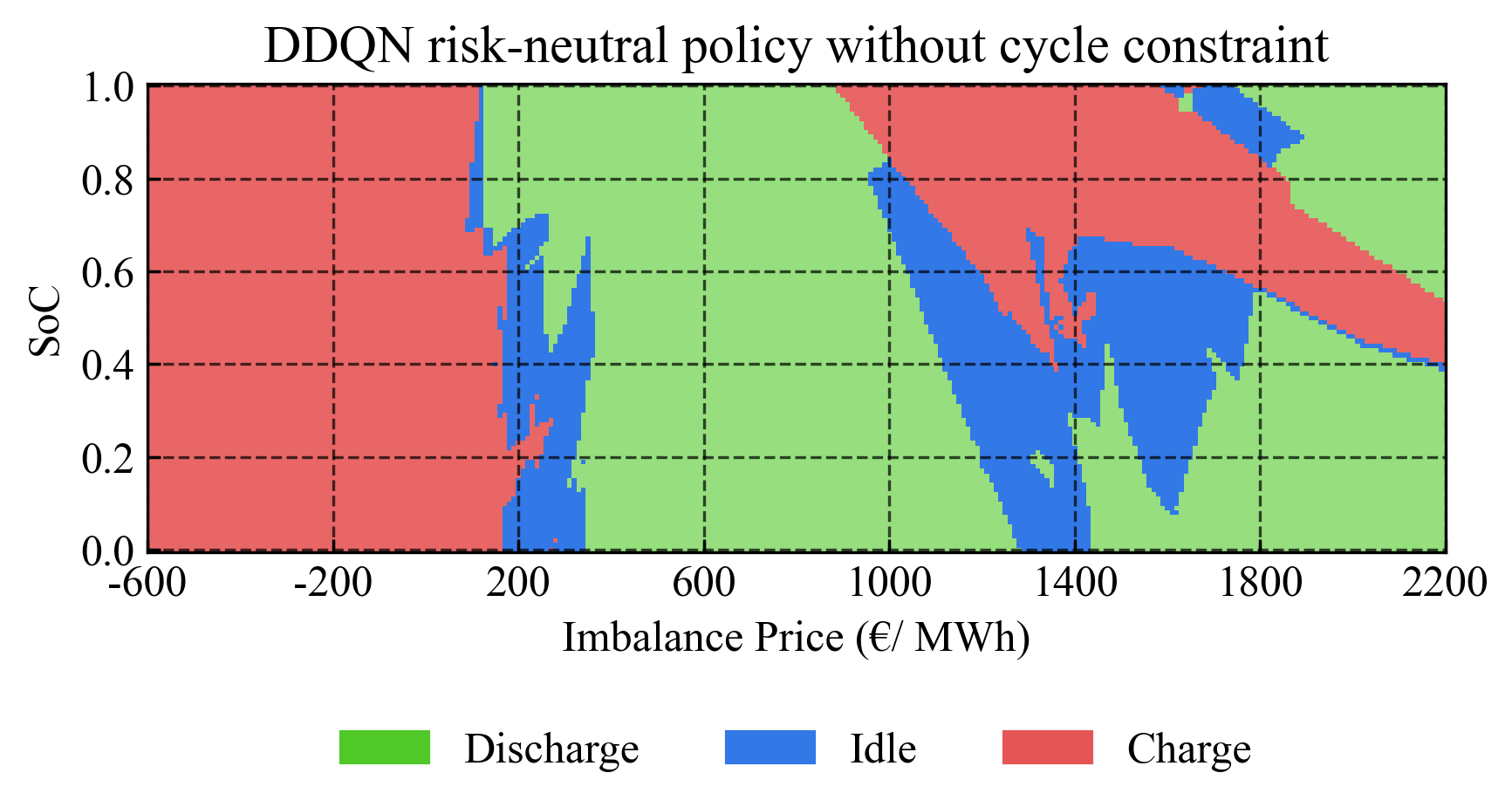}
        \caption{}
        \label{fig:DDQN policy without cycle}
    \end{subfigure}
    \begin{subfigure}{0.49\textwidth}
        \includegraphics[width=\textwidth]{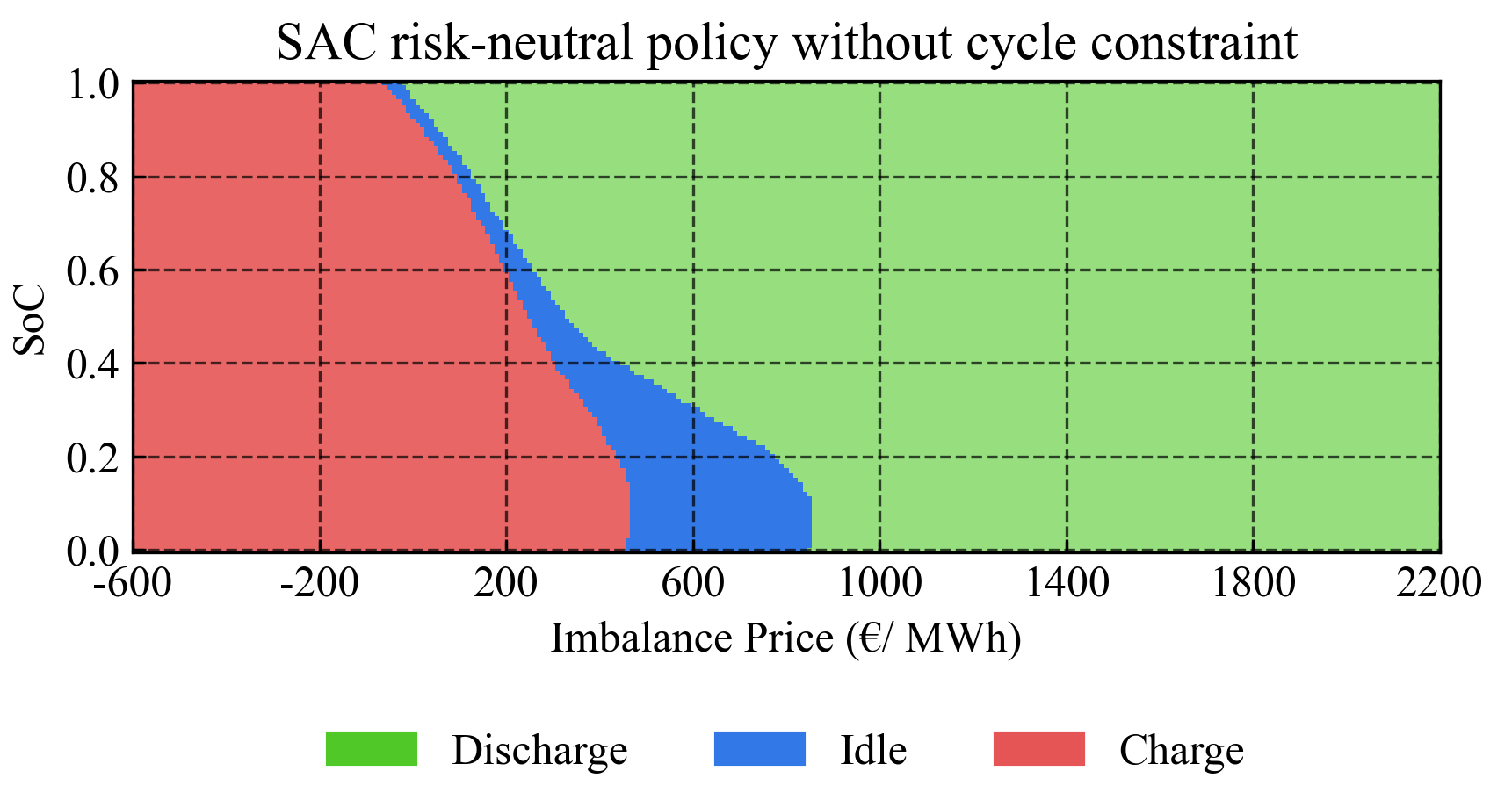}
        \caption{}
        \label{fig:SAC policy without cycle}
    \end{subfigure}
    \hfill
    \begin{subfigure}{0.49\textwidth}
        \includegraphics[width=\textwidth]{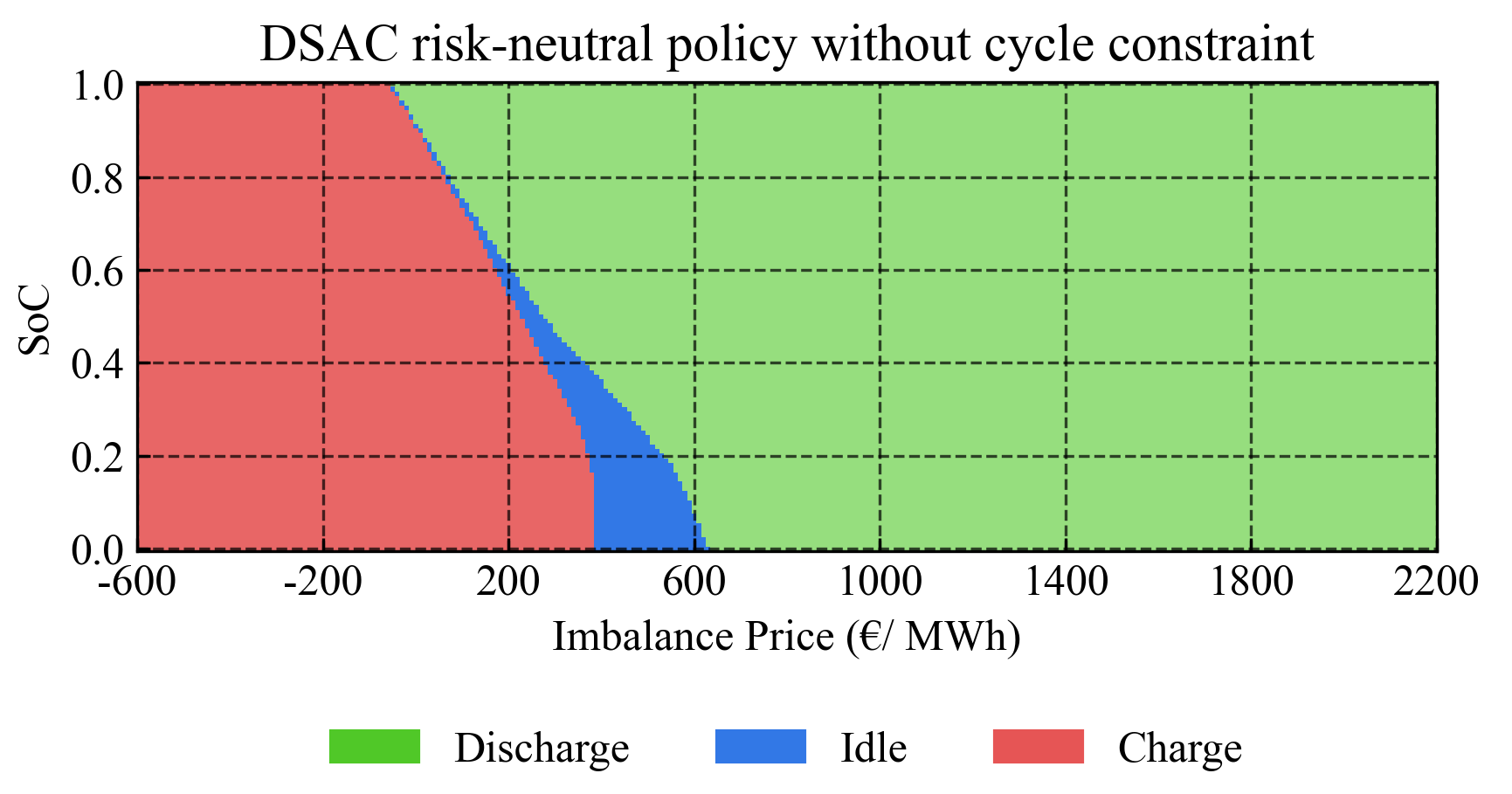}
        \caption{}
        \label{fig:DSAC policy without cycle}
    \end{subfigure}
    \caption{The projection of the learned policy in the risk-neutral without cycle constraint scenario for (a)~DQN, (b)~DDQN, (c)~SAC, and (d)~DSAC.}
    \label{fig:policies without cycle}
\end{figure}

\begin{figure}[H]
    \centering
    \includegraphics[width=0.48\textwidth]{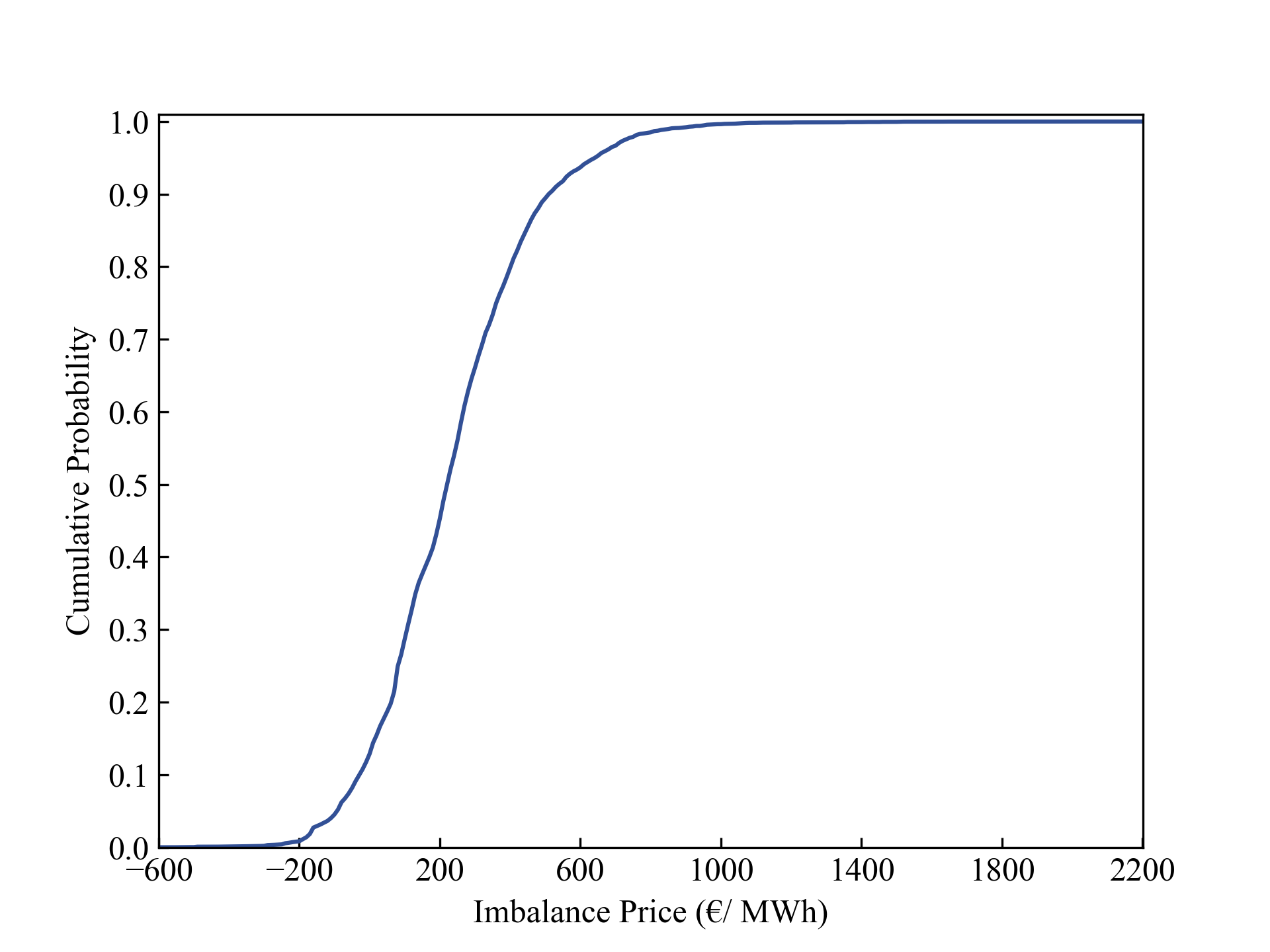}
    \caption{The cumulative distribution of the imbalance price in 2022.}
    \label{fig:Price Distribution}
\end{figure}

To analyze and study the learned policy of the four RL methods, the policy heatmaps are illustrated in \cref{fig:policies without cycle}. Since SoC and forecasted imbalance price are the two most determinative features for the agent, we show the learned policy with respect to these two input features, which are also informative to interpret the policy. \Cref{fig:policies without cycle} shows that the SAC and DSAC methods can learn a more meaningful and smooth policy compared to the DQN and DDQN methods. For DQN and DDQN, the Q-value function overestimates the value of rarely seen states and out-of-distribution (OOD) actions in these rare sates due to the max operator and the reliance of the estimated Q values on inputs from the same distribution as its training set. This overestimation results in policies that choose OOD actions. According to \cref{fig:Price Distribution}, the forecasted imbalance price rarely goes beyond 850 \texteuro/ MWh (the probability is 1\%). It means that the DQN and DDQN methods overestimate Q values for this area and take OOD actions. \Cref{fig:policies without cycle,,fig:Price Distribution} reveal some correlation between the learned policy by DSAC and the price distribution. The agent always charges the BESS when the price is within the lower 7\% quantile (lower than $-$60 \texteuro/ MWh), regardless of the SoC level. The agent never takes the charging action for the 25\% highest prices (prices higher than 380 \texteuro/ MWh), even if the BESS is empty. The BESS is always discharged when the price lies in the upper 5\% quantile (higher than 640 \texteuro/ MWh). For the median price (roughly 220 \texteuro/ MWh), the BESS is discharged if the SoC is higher than 60\%, does nothing when the SoC is between 60\% and 50\%, and is charged if the SoC is lower than 50\%. Generally, the agent learns a milder slope boundary for the discharge action. If the BESS with a low SoC level is discharged, the agent needs to quickly recharge the BESS to make sure it can still make money. This quick recharging increases the risk of charging at a higher price. Therefore, by decreasing the SoC, the area of idle action becomes larger.

\subsection{Arbitrage Strategy with Cycle Constraint (Q2)}
\begin{figure}[h]
    \centering
    \begin{subfigure}{0.49\textwidth}
        \includegraphics[width=\textwidth]{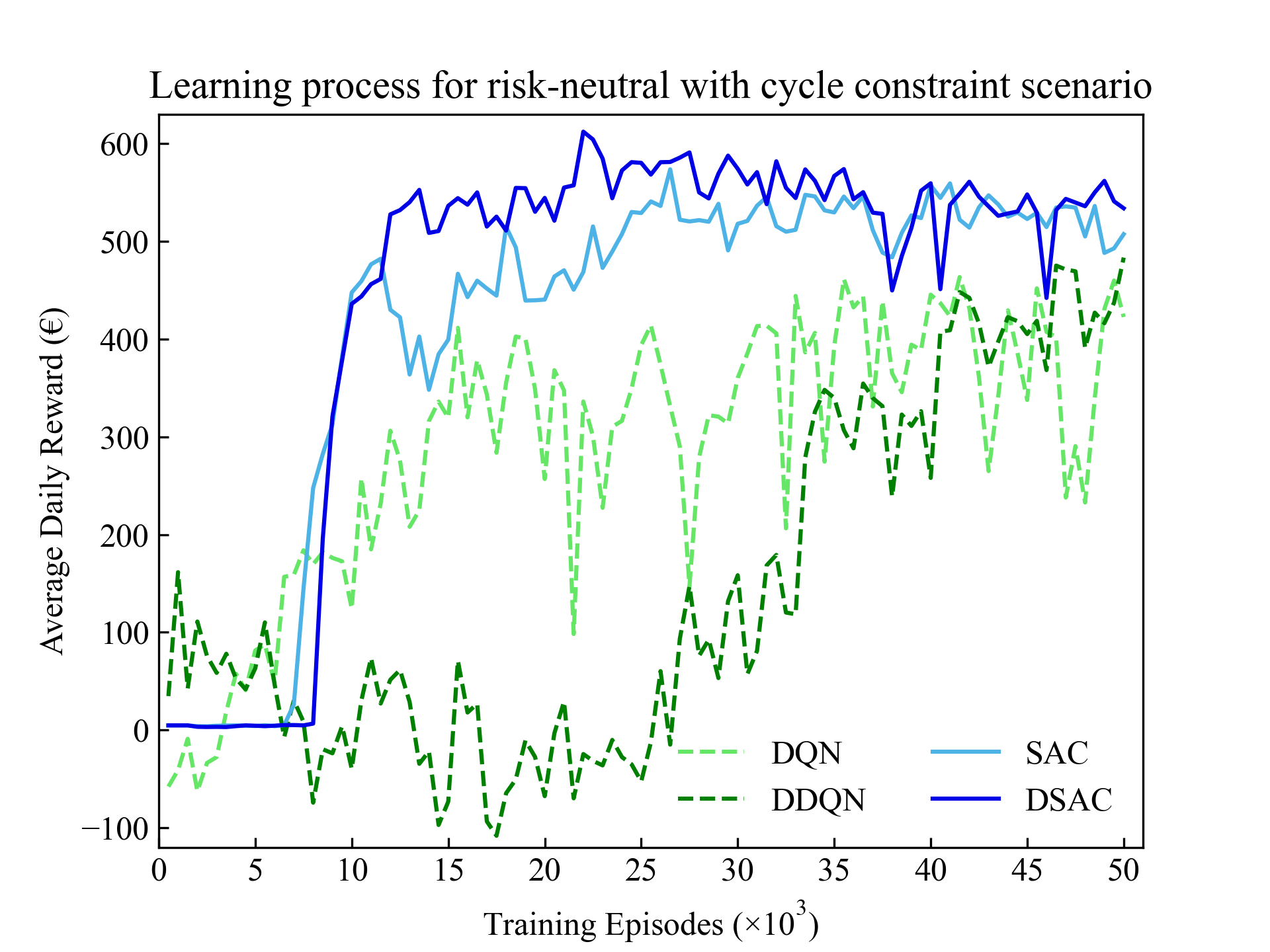}
        \caption{}
        \label{fig:learning process with cycle_reward}
    \end{subfigure}
    \hfill
    \begin{subfigure}{0.49\textwidth}
        \includegraphics[width=\textwidth]{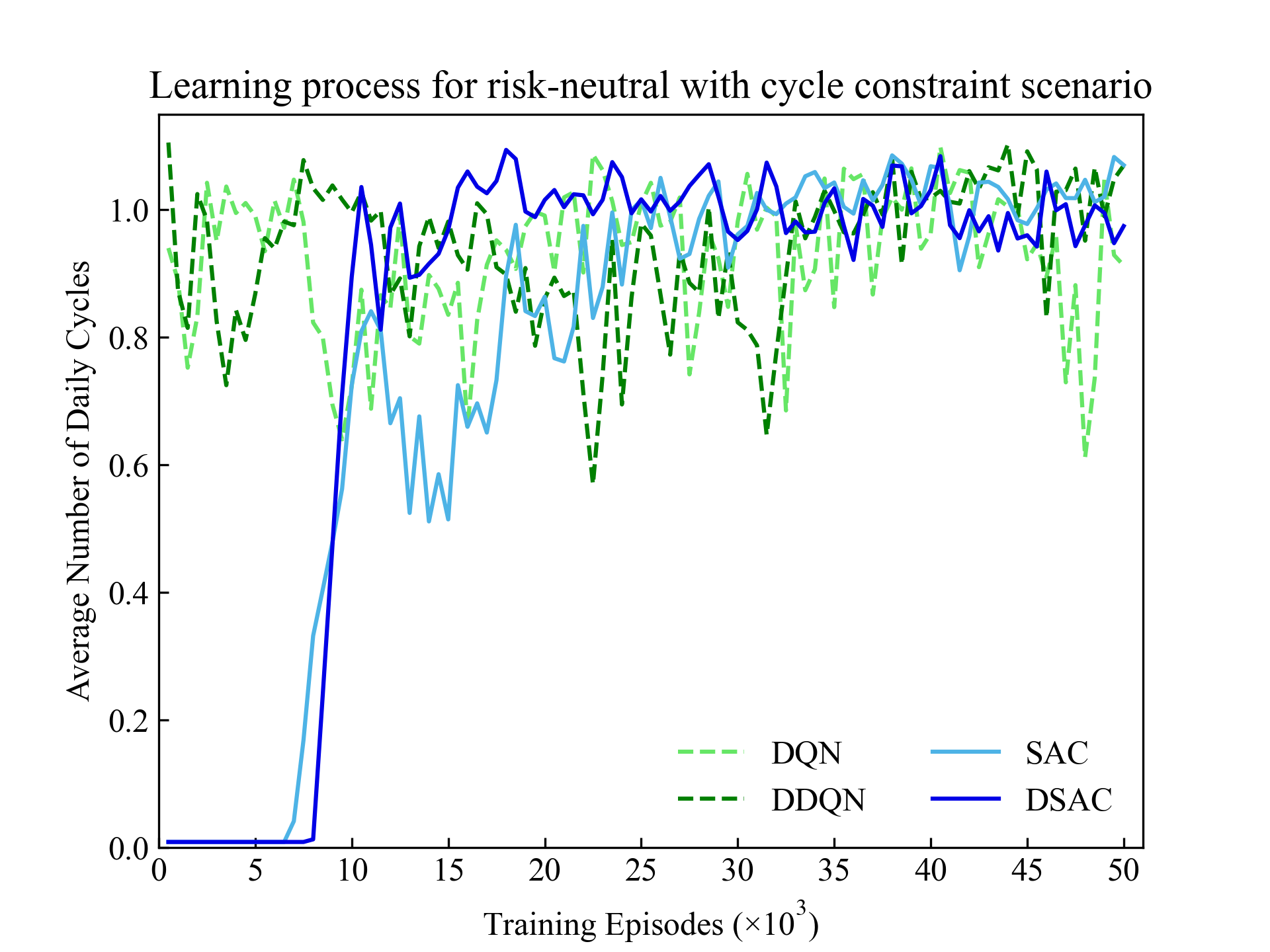}
        \caption{}
        \label{fig:learning process with cycle_cycle}
    \end{subfigure}
    \caption{The learning process of the four RL methods for the risk-neutral with cycle constraint scenario, in terms of (a)~the average daily profit, and (b)~the average daily number of cycles.}
    \label{fig:learning process with cycle}
\end{figure}

\Cref{fig:learning process with cycle} shows the learning process of the RL methods for the risk-neutral scenario when the limitation is applied to the daily number of cycles. Similar to the previous scenario, the DSAC method surpasses other methods by converging to a higher reward with a fewer number of cycles. According to \cref{tab:risk-neutral results}, although the average daily profit of the DSAC method is less than that of the SAC method, the DSAC method earns this profit by consuming fewer number of cycles. In other words, the DSAC method achieves a 14.6\% improvement in the proportional reward per cycle compared to the SAC method.
Furthermore, the SAC and DSAC methods converge faster than the DQN and DDQN methods due to their efficient exploration. Since in DQN and DDQN the learned policy is deterministic, the $\epsilon$-greedy exploration technique needs to be used. On the other hand, the SAC and DSAC methods learn a stochastic policy and use the learned probabilities for exploration. Thus, instead of always considering a fixed exploration probability of $\epsilon$ for all states, the probability of exploration depends on the current state. For a given state, when the probability of one action is close to 1, the agent almost always exploits and hardly explores. Conversely, when probabilities of all actions are close to each other, the agent most of the time explores to find the best action for that state. Consequently, the SAC and DSAC methods are more data efficient than the DQN and DDQN methods.

\begin{figure}[h]
    \centering
    \includegraphics[scale=0.23]{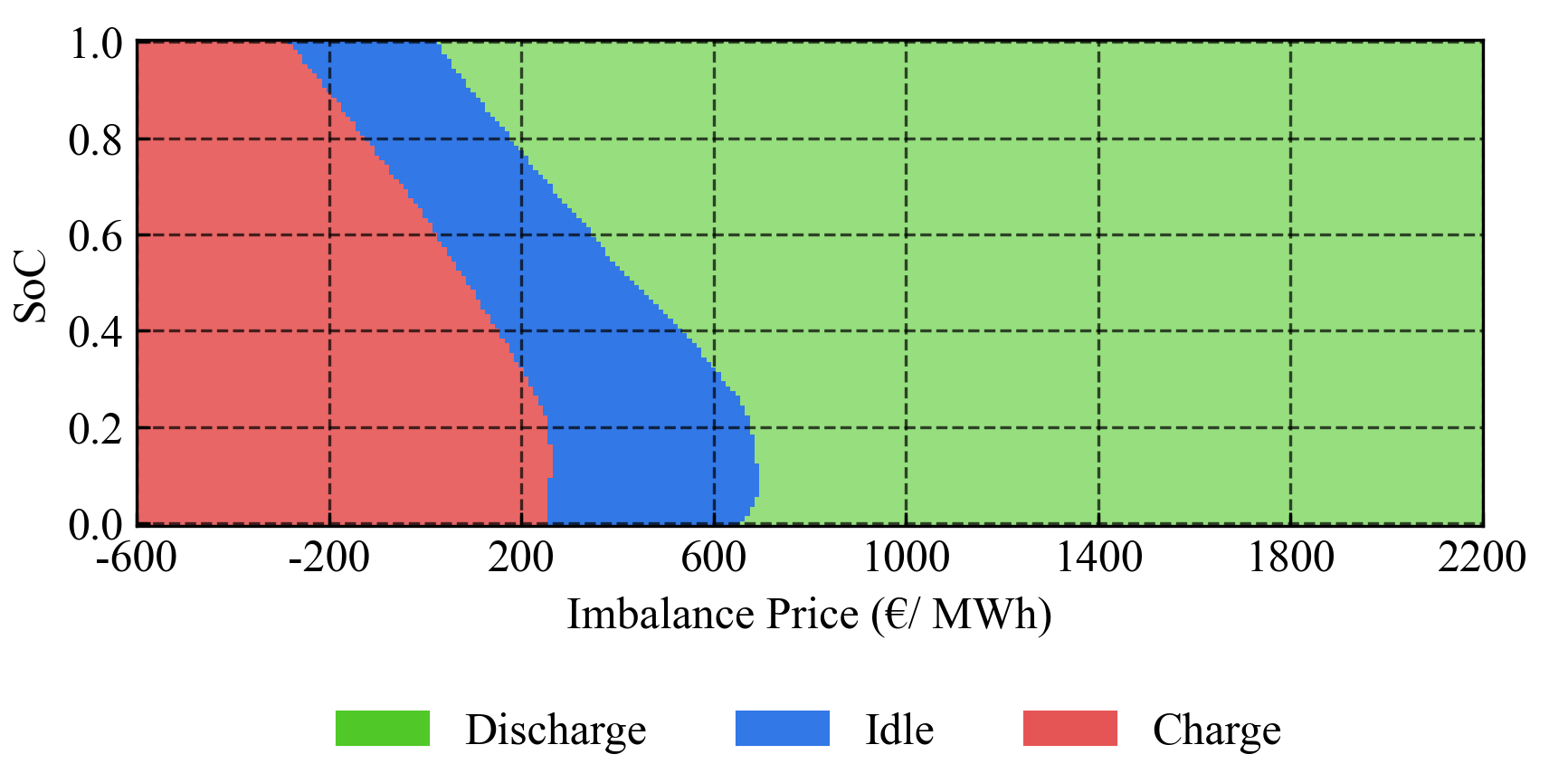}
    \caption{The projection of the learned policy in the risk-neutral with cycle constraint scenario for DSAC.}
    \label{fig:policy with cycle}
\end{figure}

\begin{figure}[h]
    \centering
    \begin{subfigure}{0.49\textwidth}
        \includegraphics[width=\textwidth]{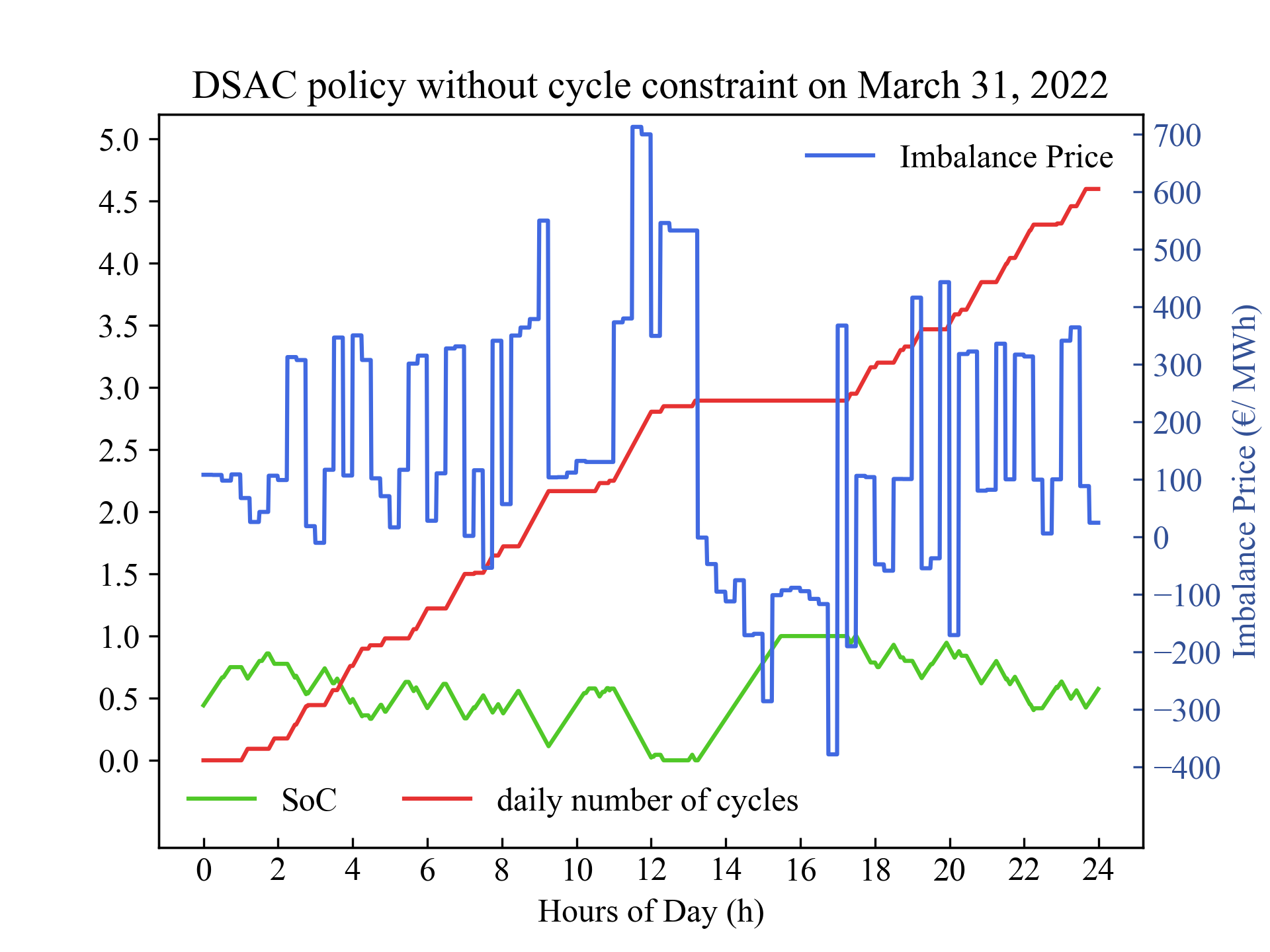}
        \caption{}
        \label{fig:policy for a day_without cycle}
    \end{subfigure}
    \hfill
    \begin{subfigure}{0.49\textwidth}
        \includegraphics[width=\textwidth]{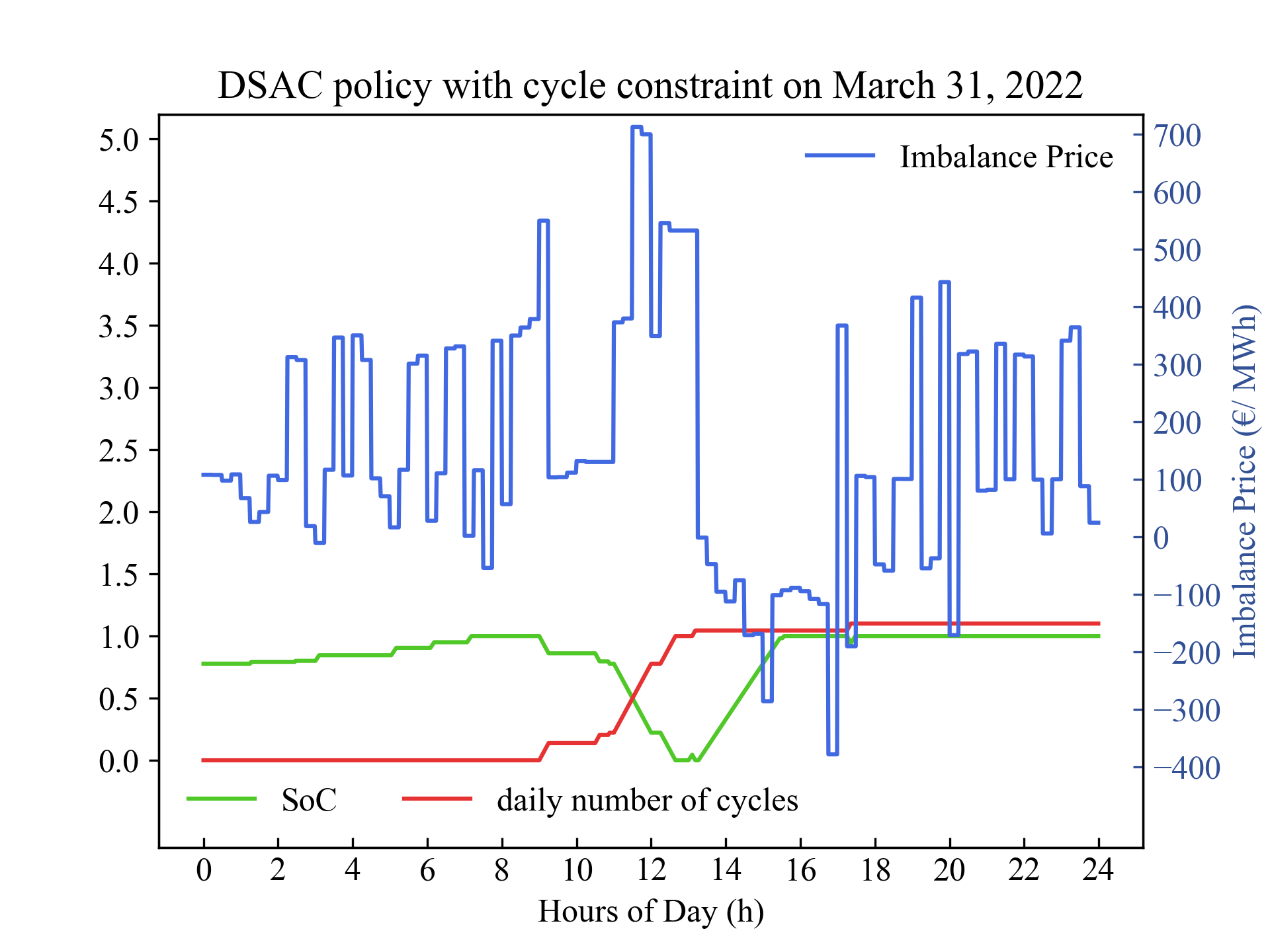}
        \caption{}
        \label{fig:policy for a day_without cycle}
    \end{subfigure}
    \caption{The performance of the trained agent by the DSAC method on March 31, 2022 (a)~without and (b)~with considering cycle constraint.}
    \label{fig:policy for a day}
\end{figure}

The learned policy of DSAC when considering the cycle constraint is illustrated in \cref{fig:policy with cycle}.
Note that the displayed policy is a projection of the learned policy, as the learned policy depends on more than two features and thus is more complicated than the figures shown.
The logic behind the learned policy with and without the cycle constraint consideration, which is charging at cheap prices and discharging at expensive prices, is nearly identical.
The main difference between these learned policies is in the size of the idle action area.
Adding the cycle constraint makes the agent more conservative and increases the idle action area.
Moreover, by limiting the number of cycles, the agent recharges the BESS less frequently due to reduced discharging.
As a result, in this scenario, the agent recharges the BESS at cheaper prices compared to the previous scenario.
To show the performance of the learned DSAC agents in a real-life case, the learned agents are tested using data from March 31, 2022.
As \cref{fig:policy for a day} shows, there is one major peak in the imbalance price from 11:00 to 13:15 and one major valley from 13:30 to 17:00 on this day.
Both agents properly respond to these prices: the agent without the cycle constraint reacts to roughly all fluctuations in the imbalance price, even small ones (such as the price fluctuation between 4:30 and 6:00, or between 20 and 21:30).
However, another agent mostly focuses on more significant fluctuations to limit the number of charging cycles.

\subsection{Arbitrage Strategy with Risk Management (Q3)}
\begin{figure}[h]
    \centering
    \includegraphics[width=0.5\textwidth]{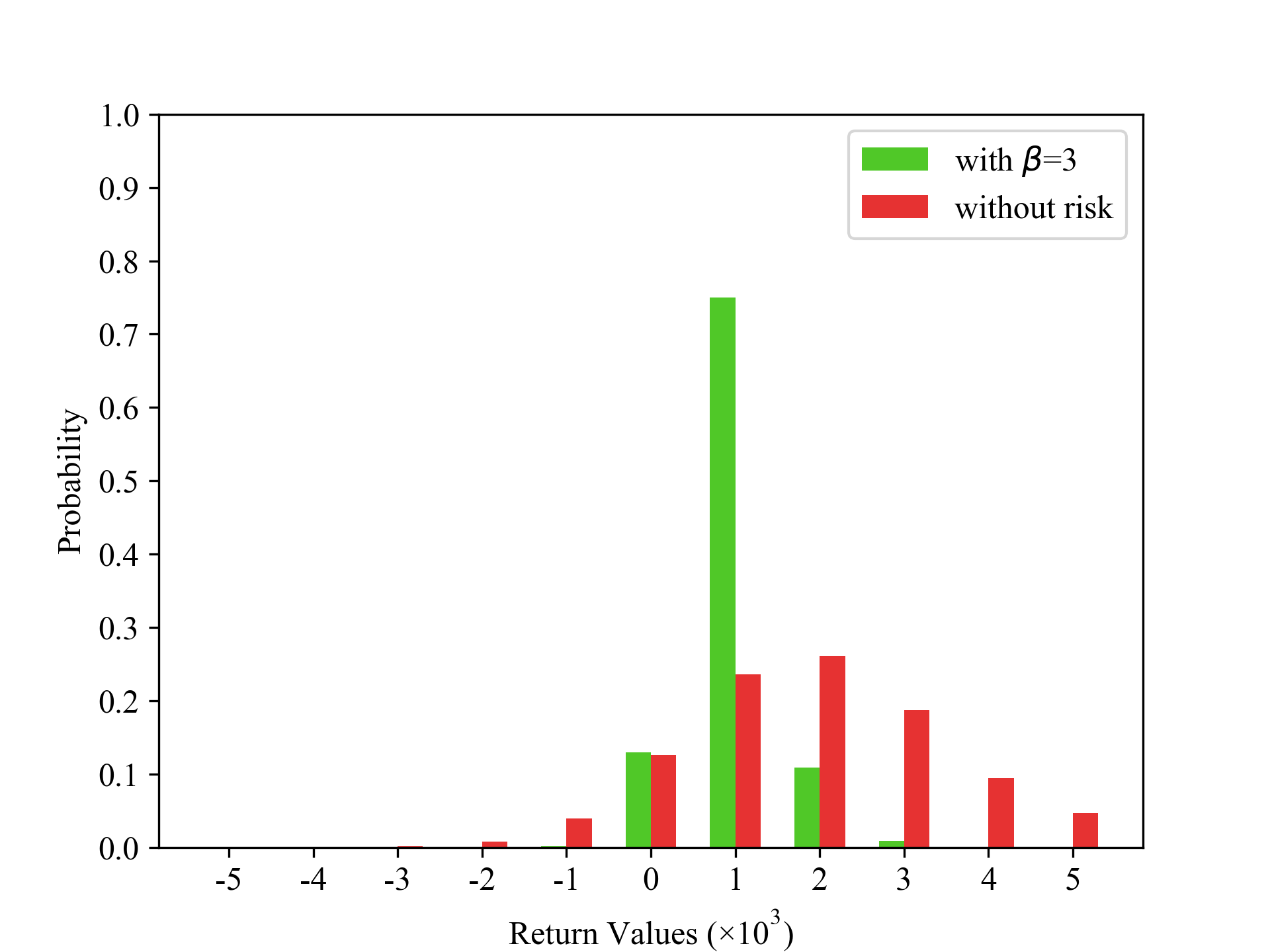}
    \caption{The learned critic network for the risk-neutral ($\beta$ = 0) and risk-averse ($\beta$ = 3) agents.}
    \label{fig:Z function}
\end{figure}

\begin{figure}[h]
    \centering
    \includegraphics[scale=0.49]{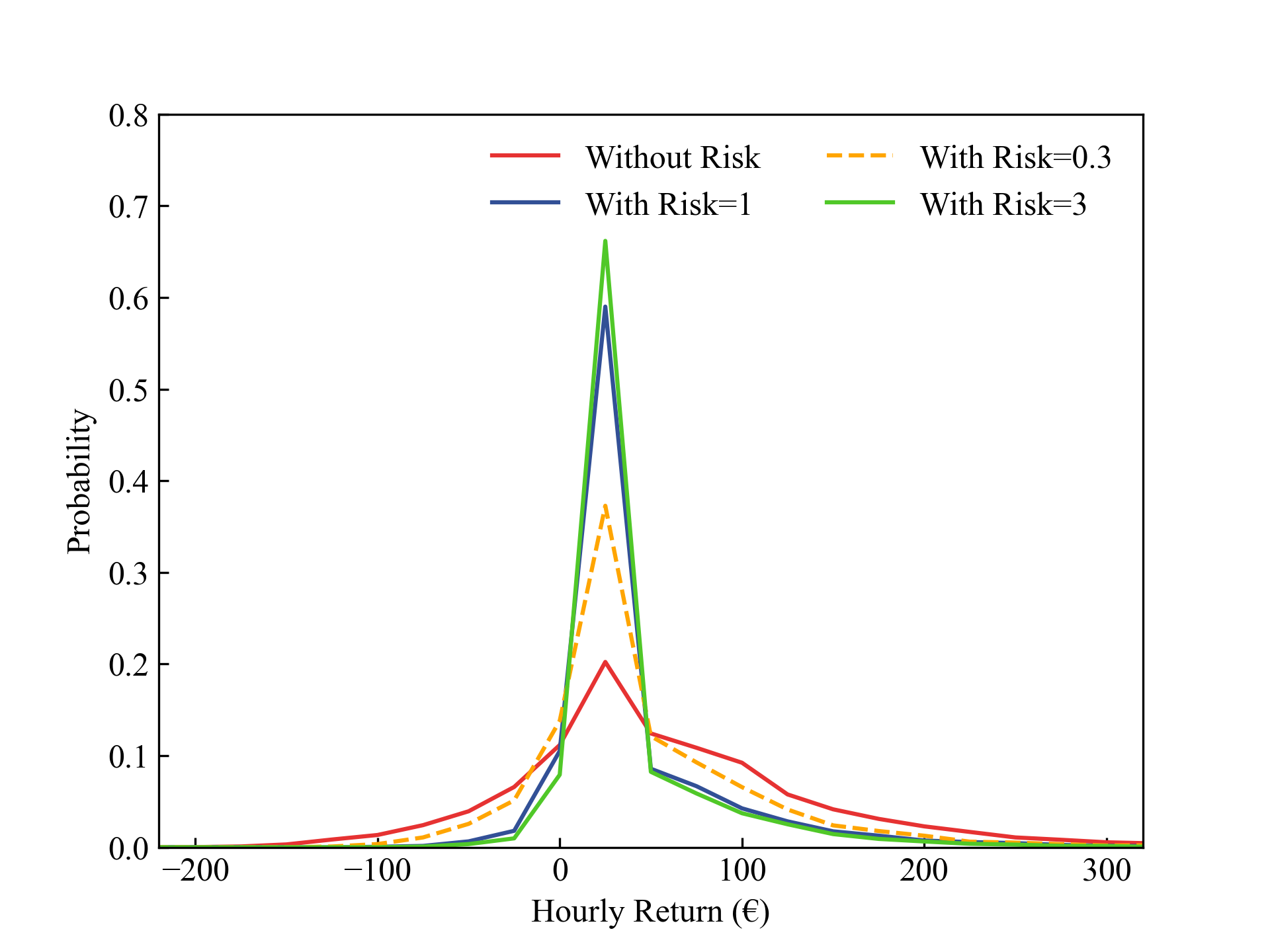}
    \caption{The probability distribution of hourly profit with and without the risk.}
    \label{fig:Probability Distribution of Hourly Return}
\end{figure}

To answer Q3, we train the DSAC agent without the cycle constraint consideration for varying $\beta$ values. Results in \cref{tab:risk-sensitive results} shows that the risk-averse agent with $\beta=3$ experiences a 54.8\% reduction in the average daily profit compared to the risk-neutral agent, but given that it avoids risky behavior, we note a higher profit per cycle. \Cref{fig:Z function} illustrates the difference between the learned critic network for the fully risk-averse and risk-neutral agents. The learned critic network for the fully risk-averse agent is narrower due to applying the risk measure function (VaR) instead of the expectation. Also the VaR values align with this observation: VaR values for the risk-neutral and fully risk-averse critic networks are equal to $-$589.2\texteuro~and $-$240.5\texteuro, respectively. The probability distribution of the hourly profit for test data is shown in \cref{fig:Probability Distribution of Hourly Return}.
Based on \cref{fig:Probability Distribution of Hourly Return}, the risk-averse agent successfully hedges against the uncertainty in the imbalance price and mitigates the tail of the hourly profit distribution.\footnote{Note that both the left- and right-tails are reduced, although from the risk perspective especially the lower (negative) return values should be avoided.}
The VaR value of each distribution is provided in \cref{tab:risk-sensitive results}.

\begin{table}[H]
    \centering
    \caption{Evaluation of DSAC method on the test set in the risk-sensitive scenario ($\beta$ = 3).}
    \begin{tabular}{ccccc}
        \toprule
        Risk aversion & \begin{tabular}{@{}c@{}} Profit \\ (\texteuro/ day)\end{tabular} & \begin{tabular}{@{}c@{}} Cycles \\ (per day)\end{tabular} & \begin{tabular}{@{}c@{}} Proportional profit \\ (\texteuro/ cycle)\end{tabular} & VaR value \\
        \midrule
        $\beta=0$ & 1148.5 & 3.6 & 314.1 & $-$71 \\
        $\beta=0.3$ & 796.7 & 2 & 399 & $-$48.5 \\
        $\beta=1$ & 593.9 & 1.25 & 474.6 & $-$32.5 \\
        $\beta=3$ & 518.9 & 1 & 518.9 & $-$24.7 \\
        \bottomrule
    \end{tabular}
    \label{tab:risk-sensitive results}
\end{table}

\begin{figure}[h]
    \centering
    \includegraphics[scale=0.23]{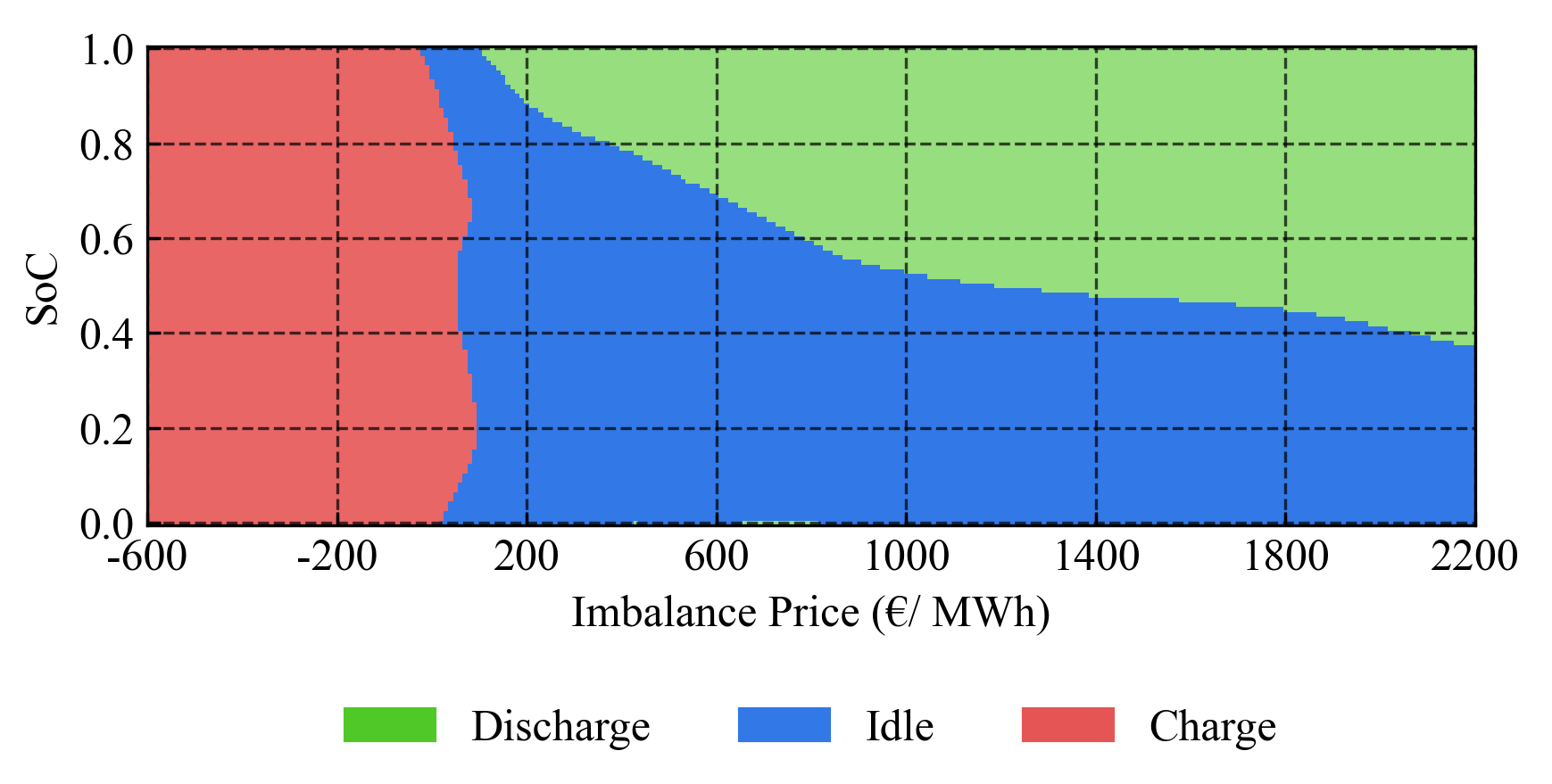}
    \caption{The projection of the learned policy using DSAC for the risk-averse agent}
    \label{fig:policies with risk}
\end{figure}

\begin{figure}[h]
    \centering
    \includegraphics[scale=0.5]{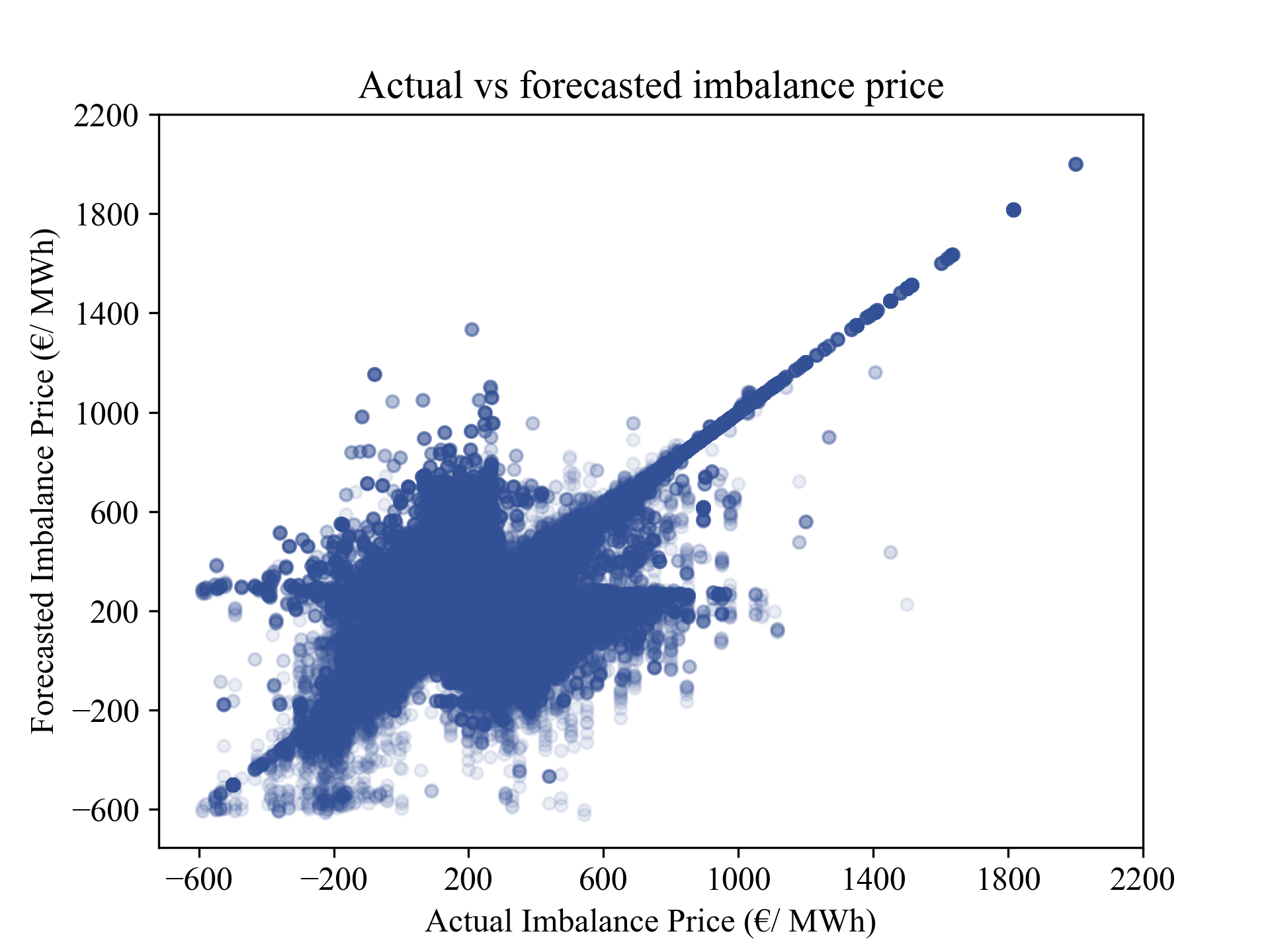}
    \caption{Actual imbalance price vs. forecasted imbalance price}
    \label{fig:actual vs forecasted price}
\end{figure}

\Cref{fig:policies with risk} shows the learned risk-averse policy when $\beta=3$. Compared to \cref{fig:policies without cycle}, we note that the idle area gets significantly larger: the agent does not discharge the battery when the SoC is low. In this way, the agent makes sure that the battery has always enough energy to inject into the grid when the price is high. Moreover, there is an observable change in the charge threshold that can be justified by \cref{fig:actual vs forecasted price}. The charge threshold for the risk-neutral agent ranges between 0 and 400 \texteuro/ MWh. However, \cref{fig:actual vs forecasted price} indicates that within this range, the actual price is significantly uncertain and the chance of charging battery at a price larger than the forecasted value is high. To mitigate this risk, the risk-averse agent learns a lower charge threshold. The risk-averse agent charges the battery at cheaper prices to minimize the risk of charging at a high price resulting from inaccurate price predictions.

\section{Conclusion}
\label{sec:Conclusion}
In this paper, a DRL-based control framework for BESS was proposed to obtain energy arbitrage strategies in the imbalance settlement mechanism. In the proposed control framework, in addition to considering a constraint on the daily number of cycles, the degree of risk taking in the learned arbitrage strategy can be adjusted based on the risk preference of BRPs. To evaluate the performance of the proposed control framework, two state-of-the-art RL methods, i.e., DQN and SAC, and their distributional variants have been implemented. The results for the Belgian imbalance price in 2022 showed that the DSAC method outperforms other methods (i.e., both the non-distributional baselines as well as DDQN) in all experiments. DSAC improves the average daily profit in the experiment without cycle constraint by 53.1\% and in the experiment with cycle constraint by 43.9\%, respectively, compared to the (worst performing) DQN method. The dominance of SAC over DQN in terms of data efficiency and mitigating Q-value overestimation, stem from replacing the max operator in the Bellman equation with the expectation operator. Moreover, the distributional methods exhibit better performance than the standard RL methods because they estimate the full probability distribution of returns rather than the expectation of returns, and they resolve instability in the Bellman optimality operator. 

In a first experiment, without considering cycle constraints, we noted that the DSAC agent learned a smooth and rational policy: it learned to charge the battery when the price is very cheap (within the lower 7\% quantile), discharge when the price is very expensive (within the upper 5\% quantile), and take the action based on the SoC for prices in between. In a second experiment, including the cycle constraints, the cycle-aware arbitrage strategy expectedly showed a larger `idle' action area compared to the case without cycle constraints, effectively leading to a lower number of cycles used. The trained cycle-aware agent tended to respond only to major peaks and valleys in the imbalance price due to the limited number of cycles, while the cycle-unaware agent reacted to almost all fluctuations in the imbalance price. Our study of risk-sensitive agents showed that the risk-averse arbitrage strategies make the distribution of hourly profit narrower and mitigate the tail of the distribution. Indeed, the risk-averse agent charges the battery at lower prices to mitigate the risk associated with inaccurate price forecasts and avoid incurring higher charging costs.

Finally, we note that in this paper, the day-ahead schedule for the battery was set to zero. In future research, the proposed control framework will be generalized by taking into account energy arbitrage between the day-ahead market and the imbalance settlement mechanism. Studying the effect of considering a continuous action space instead of a discrete one forms another next step to take.

\appendix

\section{Comparing DQN with FQI}
\begin{figure}[h]
    \centering
    \includegraphics[scale=0.49]{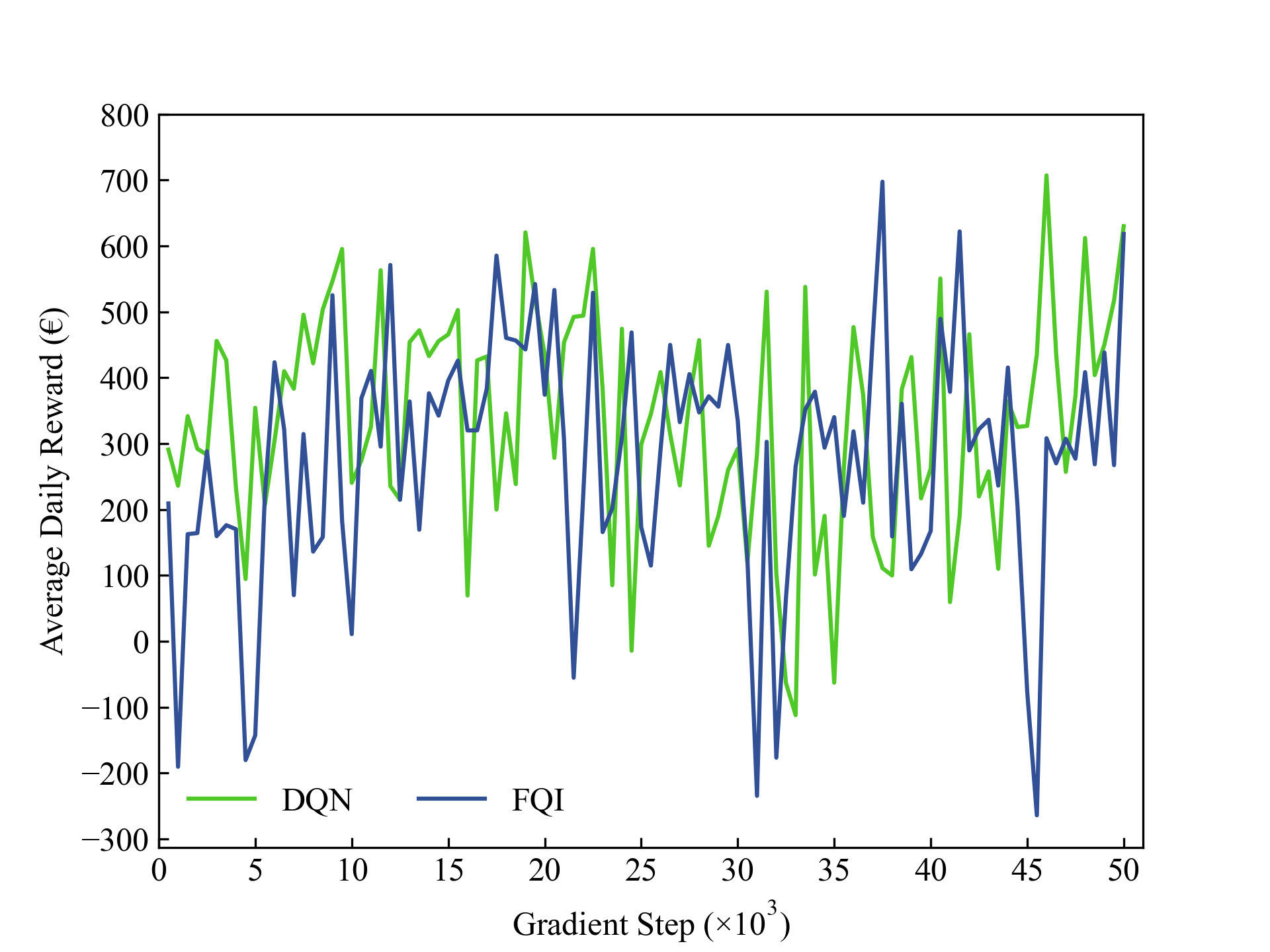}
    \caption{The learning process of the DQN and FQI methods for the small experiment.}
    \label{fig:learning process for small experiment}
\end{figure}

The FQI method~\cite{riedmiller2005} is another widely used value-based method. In~\cite{Lago2021controller}, FQI is used to obtain a 15-minute-based arbitrage strategy in the imbalance settlement mechanism. In this section, a small experiment is carried out to compare the performance of the DQN and FQI methods. In this experiment, the methods are trained on the first nine days of February and evaluated on February 10, 2022. The architecture of the neural network used in the FQI method is the same as that of the DQN method. The experience replay buffer size, number of iterations, and number of episodes are \num{16384}, 400, and 500, respectively. In accordance with \cref{fig:learning process for small experiment}, both methods perform almost similarly. However, the run time of the FQI method is roughly 5 times greater than that of the DQN method and even gets worse by increasing the experience replay size and the number of episodes. The reason for the longer run time for FQI is its number of iterations: in each episode, the Q network is trained for the mentioned number of iterations. Thus, the FQI method is inappropriate for obtaining the arbitrage strategy.

\printbibliography

\end{document}